\pdfoutput=1
\documentclass[11pt]{article}

\usepackage[final]{acl}
\usepackage[T1]{fontenc}
\usepackage[utf8]{inputenc}
\usepackage[nopatch=footnote]{microtype}
\usepackage{times}
\usepackage{inconsolata}
\usepackage{graphicx}
\usepackage{amsmath}
\usepackage{booktabs}

\usepackage{listings}
\usepackage{xcolor}

\usepackage{pdflscape}

\usepackage{longtable}   % for tables that span pages
\usepackage{pdflscape}   % for landscape pages

\usepackage{latexsym}
\usepackage{tipa}
\usepackage{tikz}
\usepackage{makecell}
\usepackage{rotating}
\usepackage{amssymb}
\usepackage{pifont} 
\usepackage{tabularray} 
\usepackage{array}
\usepackage{fdsymbol}
\usepackage{cleveref}
\usepackage{multirow}
\usepackage[scaled=0.85]{helvet}
\usepackage{tcolorbox}
\usepackage{xcolor}
\usepackage{graphicx}
\usepackage{makecell}

\usepackage{expex}

\usepackage{tabularx}

\makeatletter
\DeclareRobustCommand*{\escapeus}[1]{%
  \begingroup\@activeus\scantokens{#1\endinput}\endgroup}
\begingroup\lccode`\~=`\_\relax
\lowercase{\endgroup\def\@activeus{\catcode`\_=\active \let~\_}}
\makeatother
\newcommand{\myemph}[1]{\textsf{{\escapeus{#1}}}}

\definecolor{cambridgeblue}{RGB}{163,193,173}
\definecolor{customyellow}{HTML}{E0A000} % FECE62
\definecolor{cambridgewarmblue}{RGB}{0,189,182}

\usepackage{tabularray}
\usepackage{cellspace}
\newcommand\cincludegraphics[2][]{\raisebox{-0.3\height}{\includegraphics[#1]{#2}}}
\usepackage{setspace}
\usepackage{twemojis}

\newtcbox{\characterhighlight}{on line, colback=cambridgeblue!50, boxrule=0.2mm, left=0.5mm, right=0.2mm, top=0.2mm, bottom=0.2mm}

\newtcbox{\spacehighlight}{on line, colback=customyellow!50, boxrule=0.2mm, left=0.5mm, right=0.2mm, top=0.2mm, bottom=0.2mm}

\newtcbox{\phonemehighlight}{on line, colback=cambridgewarmblue!50, boxrule=0.2mm, left=0.5mm, right=0.2mm, top=0.2mm, bottom=0.2mm}

\usepackage{xspace}
\newcommand{\name}{\textsc{ContingentChat}\xspace}
\title{Teacher Demonstrations in a BabyLM's Zone of Proximal Development for Contingent Multi-Turn Interaction}

\author{
    {\bf Suchir Salhan}\thanks{   \textbf{ Corresponding Author:} \texttt{sas245@cam.ac.uk}}~\texttwemoji{baby_bottle}\texttwemoji{person_feeding_baby} \quad
    {\bf Hongyi Gu}~\texttwemoji{hatching_chick} \quad %\\
    {\bf Donya Rooein}~\texttwemoji{baby_chick} \quad
    {\bf Diana Galvan-Sosa}~\texttwemoji{baby_bottle}\texttwemoji{person_feeding_baby} \quad
    \\
    {\bf Gabrielle Gaudeau}~\texttwemoji{baby_bottle}\texttwemoji{person_feeding_baby} \quad
    {\bf Andrew Caines}~\texttwemoji{baby_bottle}\texttwemoji{person_feeding_baby} \quad
    {\bf Zheng Yuan}~\texttwemoji{hatched_chick} \quad
    {\bf Paula Buttery}~\texttwemoji{baby_bottle}\texttwemoji{person_feeding_baby} \\
\texttwemoji{baby_bottle}\texttwemoji{person_feeding_baby}  ALTA Institute, Dept. of Computer Science \& Technology, Cambridge University \\
\texttwemoji{hatching_chick} NetMind.AI 
\texttwemoji{baby_chick} Bocconi University 
\texttwemoji{hatched_chick} Sheffield University%}
}

\begin{document}
\maketitle
\begin{abstract}
Multi-Turn dialogues between a child and caregiver are characterised by a property called \textsc{contingency} – prompt, direct and meaningful exchanges between interlocuters.  We introduce \name, a Teacher–Student framework that benchmarks and improves multi-turn contingency in a BabyLM trained on 100M words. Using a novel alignment dataset for post-training, BabyLM generates responses that are more grammatical and cohesive. Experiments with adaptive teacher decoding strategies show limited additional gains. \name highlights the positive benefits of targeted post-training on dialogue quality and indicates that \textsc{contingency} remains a challenging goal for BabyLMs. 
\end{abstract}
\noindent

\begin{tblr}{colspec = {Q[c,m]|X[l,m]}, stretch = 0}
    % First row: Hugging Face logo + link/text
    \cincludegraphics[width=1.35em, keepaspectratio]{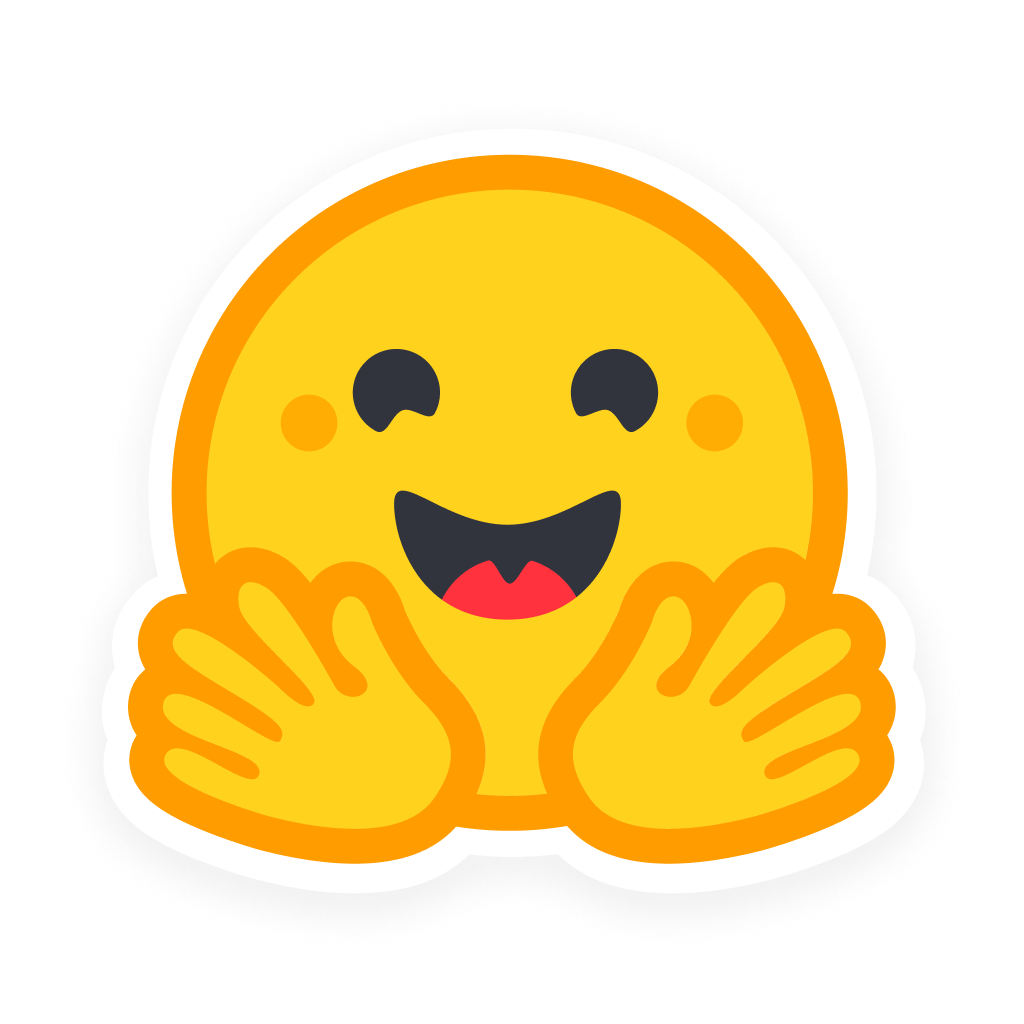} &
    \setstretch{.5}{\small\myemph{\textbf{ContingentChat} on \href{https://huggingface.co/Contingent-Chat}{HuggingFace} (Models, Tokenizers and ContingentChat Post-Training Dataset)}} \\

    % Second row: GitHub logo + link/text
    \cincludegraphics[width=1.1em, keepaspectratio]{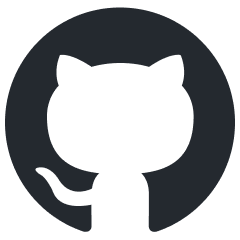} &
    \setstretch{.5}{\small\myemph{Training, Post-Training \& Analysis Code Open-Sourced on \href{https://github.com/suchirsalhan/babylm-interaction}{GitHub}}}
\end{tblr}

\section{Introduction}

%Interaction between a child and a caregiver is an important source of input in First Language Acquisition (L1).
Conversational interaction with caregivers is crucial for children learning their first language (L1) or first languages (L1s).
Linguistic interaction provides a source of primary linguistic data (PLD) for the learner, supporting the acquisition of formal competence of the target L1 grammar. It also serves as input for the acquisition of functional and pragmatic competence in the L1.
%In particular, it allows a learner to acquire the qualities of prompt and meaningful interaction in fluent and connected communication. 
%This concept is closely related to cohesion and coherence in text. 
A key feature of child-caregiver conversations to promote language learning is \textsc{contingency}. Contingent interactions are the prompt and meaningful exchanges between a caregiver and infant that form the foundation for fluent and connected communication \cite{masek2021language}. 

In this paper, we draw upon the notion of contingency in the context of cognitively-inspired small language modelling to design \name, a 
%\emph{cognitively-inspired post-training framework to enhance the Contingency of BabyLM text generation in Multi-Turn dialogic interaction with a Teacher LLM}.
cognitively-inspired post-training framework to enhance the contingency of BabyLM text generation in multi-turn dialogic interaction with a Teacher LLM.
\name is designed to enhance the dialogue generation capabilities of BabyLMs submitted for the \textsc{Strict} track of the BabyLM Challenge, which are trained on 100M words of developmentally-plausible training data.

\begin{figure}[t]
    \centering
    \includegraphics[width=\linewidth]{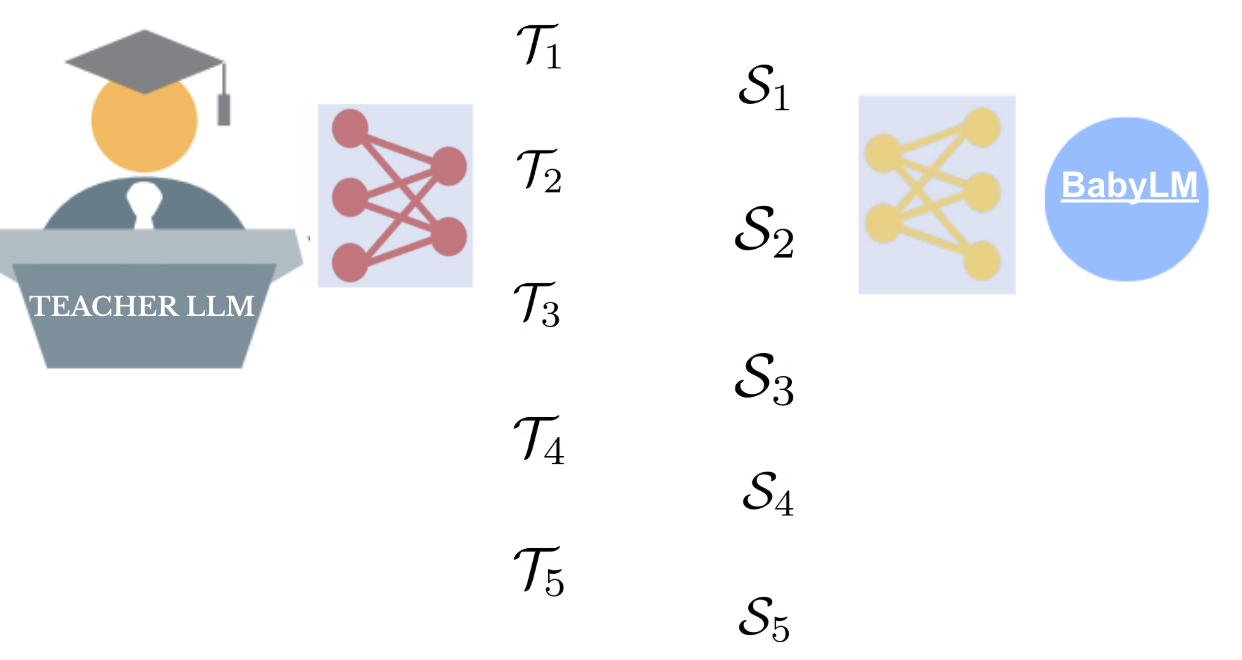}
    \caption{ We consider multi-turn dialogic interactions between a BabyLM trained on 100M words (a \textsc{Strict} model) and a Teacher LLM. The \name framework aims to improve BabyLM generations by rewarding more cohesive and coherent generations through trials-and-demonstrations in a post-training phase.}
    \label{fig:example}
    \vspace{-6mm}
\end{figure}

Previous editions of the BabyLM Challenge \citep{warstadt-etal-2023-findings, hu-etal-2024-findings} have evaluated submitted models trained with 10M (\textsc{Strict-Small}) or 100M (\textsc{Strict}) words on benchmarks of formal linguistic competence. These have included BLiMP \citep{warstadt-etal-2020-blimp-benchmark}, or BLiMP Supplement, which consist of minimal pairs 
designed to test the grammaticality judgements which language models are capable of making
%that consist of subtasks that evaluate the turn-taking capabilities of LMs and Question-Answer Congruence 
\citep{hu-etal-2024-findings}.
However, as \citet{charpentier-samuel-2024-bert} note, none of these benchmarks are well-suited for evaluating causal language model text generation, let alone evaluating generation quality or the alignment of model generations in an acquisition-inspired manner. 

\name is an iterative Teacher-Student post-training pipeline enabling interaction between a small language model (the \emph{BabyLM}) and a much larger language model (the \emph{Teacher}; see~\Cref{fig:example}). The BabyLM and Teacher LLM repeatedly interact on sub-dialogues selected from a 30M word annotated English dialogue corpus (\textbf{The \name Alignment Dataset}). These 30M words come from the Switchboard Dialog Act Corpus \citep{godfrey1992switchboard, stolcke2000dialogue}, which we have annotated with cohesion metrics from different tools such as NLTK\footnote{\url{https://www.nltk.org/}}, Spacy\footnote{\url{https://spacy.io/}}, Tools for the Automatic Analysis of Cohesion 2.0 (TAACO) \citep{crossley2019tool}, and our own bespoke processing and statistical calculations. 

Our starting point in designing \name is evaluating the dialogue generation capabilities of BabyLMs trained on 100M words of developmentally-plausible training data. We initially observed that \textsc{Strict} BabyLMs are prone to self-repetition and struggle to produce coherent responses to prompts from a wide range of Teacher LLMs. We use our initial analysis to inform the experimental design of \name. 

In our framework, a BabyLM produces a continuation in a dialogue, and the Teacher LLM improves it according to strict anti-repetition and coherence guidelines. \name rewrites BabyLM outputs to improve coherence and naturalness, which we treat as a chosen or edited response. This results in preference pairs – (1) original outputs of a BabyLM and (2) edited and improved responses by a Teacher LLM. \name accumulates these preference pairs for post-training the BabyLM to gradually, over successive training rounds, guide it towards producing high quality, contextually-appropriate responses which are closer to the teacher's.

Our preliminary experiments indicate that models trained on fewer than 100M words struggle to sustain post-training interactions in multi-turn dialogues with a Teacher LLM. Despite achieving competitive scores on the BabyLM Evaluation benchmark, these smaller models often exhibit unstable or undesirable generation behaviours—such as self-repetition or incoherent responses—during post-training. Moreover, improvements from reward-based post-training appear inconsistent, suggesting that a minimum data scale is necessary for models to effectively benefit from reward learning and exhibit genuine intrinsic improvement. 

Effective Teacher Demonstrations to a BabyLM are theoretised to be within a BabyLM’s Zone of Proximal Development (ZPD). Vygotsky's Zone of Proximal Development (ZPD) proposes that learners are capable of acquiring new knowledge with support, up to the point at which such knowledge would be too complex to acquire \citep{vygostky-1978}. Using \name, we first systematically experiment in Experiment 1 with different  post-training conditions that can potentially enhance the contingency of multi-turn BabyLM–LLM dialogues trained on 100M words. Experiment 2 assesses the benefits of using an adaptively-decoded Teacher LLM. This is motivated by one prevailing idea from language acquisition, known as the \emph{Goldilocks Principle} \citep{kidd-2014}. Children naturally focus on input that is neither too simple nor too difficult but at the right level of challenge for learning, which suggests there might be benefits of adapting Teacher turns to the observed proficiency level of the BabyLM. We find scaffolding BabyLM outputs using reward models that encourage stricter adherence to Teacher Demonstrations forms a stable Zone of Proximal Development, effectively enhancing  BabyLMs’ capacity to generate contingent, contextually grounded dialogue via constrained policy updates. 

\section{Interactive Language Learning}

\subsection{Naturalistic Interaction in First Language (L1) Acquisition}

In addition to acquiring formal competence of their first language, children have to learn to become competent conversational partners with others. This involves learning a complex set of skills. Spoken dialogues involve rapid exchanges of turns and interlocutors tend to use prediction and inference to keep a conversation flowing and coherent \citep{levinson2016turn}.

Beyond learning specific skills like turn-taking in dialogues,  contingency is a conversational behaviour that we define broadly, following \citet{masek2021language} and \citet{agrawal2024automatic}, as the ability to produce \textit{multi-turn dialogues}\footnote{Beyond the specific properties that characterise the quantity and quality of individuals language interactions, \textbf{contingency} can refer to the more general statistical learning process by which associations are formed between cues and outcomes (\citet{ellis2006language, ellis2006selective}, \citet{hsu2011probabilistic} \& \citet{guo2021language} i.a). This is a more general framing in the statistical learning literature which we apply in the more narrow domain of multi-turn dialogic interaction.}. Contingent dialogues have properties that distinguish them from successive chains of disconnected remarks or narrative monologues. Interlocutors have been theorised to operate via a Principle of Cooperativeness \citep{grice1975logic}. Grice's  Maxims of Conversation characterise some idealised characteristics of dialogue: where possible they should be informative, relevant, truthful, and clear. These maxims can be flouted or violated in adult speech for deliberate effect. More recent approaches in syntax and pragmatics have proposed varying theoretical analyses of the systematicity of dialogic interaction. \citet{wiltschko2021grammar}, for example, highlights that dialogic interaction is systematically driven by the dynamic and interactional process of finding a \textit{mutual} common ground between interlocutors during a multi-turn dialogue. \footnote{ \citet{stalnaker1978assertion},  \citet{stalnaker2002common}, 
\citet{groenendijk2009inquisitive} \& \citet{bavelas2012beyond} i.a. also offer this interpretation of interactional language in terms of Common Ground (CG).}
%Interlocutors can use particles like \texttt{`oh'} to indicate a change-of-state in interaction.

In the context of L1 acquisition, language acquisition researchers have suggested infants similarly demonstrate early systematic communicative behaviour, including via non-verbal cues like raised arms and deictic gestures
%rising intonation
\citep{heim2025rethinking}, as shown in example (1) from the Forrester Corpus \citep{forrester2002appropriating}:

\pex
\a
\begingl
\gla Ella: Whaaa↑ [raises arm] (1;00 -- Forrester Corpus) //
\endgl

\a
\begingl
\gla Ella: Yehh↑ [points to object with extended index finger] //
\endgl

\a
\begingl
\gla Father: No, what's that? Huh? //
\endgl

\a
\begingl
\gla Father: I don't know, do you? //
\endgl
\xe

\subsection{Interactive Language Learning}

Work on deep multi-agent reinforcement learning (MARL) has demonstrated that agents can acquire complex behaviours – including emergent linguistic communication \citep{lazaridou2017multi} – by repeatedly interacting with other agents in a shared learning environment. These approaches leverage co-adaptation, allowing agents to bootstrap increasingly sophisticated behaviours without requiring complex simulators or demonstration data \citep{cao2018emergent, lu2020supervised, wangtowards, lazaridou2020multi, sadler2023yes}. 

Communication between artificial agents has been useful for investigating and simulating artificial language learning experiments that can be used to explore questions about learnability and the inductive biases of neural models \citep{lian-etal-2024-nellcom, lian2025simulatingemergencedifferentialcase, kouwenhoven-etal-2024-curious, kouwenhoven-etal-2025-searching}.
%However, \citet{ter2021towards} distinguishes Interactive Language Learning as interaction between a Teacher (a caregiver role) and Student LLM (whose role resembles a child) with interaction between  them along with the environment that they share.
Meanwhile, \citet{ter2021towards} distinguishes Interactive Language Learning as interaction between a Teacher (a caregiver role) and Student LLM (whose role resembles a child) with interaction between  them along with the environment that they share.

Our framework is inspired by the findings of \citet{ma2024babysit}, whose trial-and-demonstration (TnD) learning framework showed that a student model benefits from the teacher's model choices. Their setting, however, targets word learning in language model training. \name investigates the more complex task of generating individual responses which together build a coherent dialogue.

\subsection{Zone of Proximal Development}

One property of contingent multi-turn dialogues is that they are \textit{mutual}. Caregivers are constantly adapting the contingency of outputs to keep conversations engaging during multi-turn dialogues  \citep{hallart2022contingency}. Typically-developing L1 learners exhibit delays in reaching normal adult response times when engaging in multi-turn dialogues \citep{casillas2014turn}. However, analysis of child-caregiver interactions shows that children aged between 1 and 3 years typically initiate simpler answers faster than more complex answers containing less familiar words \citep{casillas2014turn}.

One possible realisation of contingent interaction from a caregiver  might include adaptive lexical simplification to meet learner needs. This general behaviour resembles the Zone of Proximal Development (ZPD; \citet{vygostky-1978}), which in a general sense refers to the range of problems that a learner can solve with appropriate scaffolding but cannot tackle independently. \citet{cui2025investigating} apply this concept to design a curriculum for in-context learning with language models.

In our setting of Interactional Language Learning, we highlight that the ZPD is relevant in two distinct ways in contingent dialogues. Firstly, child-caregiver dialogues differ in substance between earlier and later learners, as the caregiver will estimate the ZPD of the learner. Secondly, within a multi-turn dialogue, contingency is a cycle of \textbf{anticipation} and \textbf{backtracking} as caregivers try to estimate and adaptively respond to the child's changing knowledge state and their ZPD. This results in strategies like conversational repair, including adult corrective moves which support L1 acquisition, and drawing mutual attention to form and meaning \citep{clark2020conversational}.  \citet{chouinard2003adult} link adult reformulations directly to learning outcomes. The meta-pragmatic function of communicative feedback has been emphasised by \citet{ben2021conversational} and \citet{clark2014two} who highlight how the process of interpreting and responding to feedback can help children to learn about different types of feedback. 

Other work categorises forms of repair \citep{norrick1991organization, wilkinson2011insertion, cazden2017spontaneous, agrawal2024analysing,agrawal2024automatic,agrawal2025mapping}. Feedback from caregivers to learners is neither random nor uniformly distributed across error types in L1 acquisition -- early work by \citet{hiller-fernandez-2016-data} find that certain error types, such as subject omission, attract more caregiver feedback, which has been supported by findings by \citet{nikolaus2023communicative} and \citet{nikolausthesis}. \citet{saxton2000negative} and \citet{saxton2005negative} highlight the sensitivity of child-to-adult corrections with immediate impacts of \textbf{direct negative evidence} in multi-turn dialogues and a delayed influence beyond immediate uptake.

\textsc{Self-Repair} is an important mechanism of \textit{metalinguistic awareness}, which \citet{clark2008repetition} (i.a.) have argued scaffolds L1 acquisition by constructing a shared understanding in dialogic interaction, prompted by micro-cues from caregivers (e.g., \texttt{"hm?"}) that serve as subtle dialogic devices for self-correction. \citet{forrester2008emergence} and \citet{forrester2009development} document the emergence of children’s emerging self-repair capabilities as they internalise different feedback strategies. 

\section{Evaluating BabyLMs Multi-Turn Interaction}

The aim of our proposed framework, \name, is to characterise the qualities of  contingent multi-turn dialogue between a Teacher LLM and a Student BabyLM trained on 100M words. We begin with a qualitative evaluation of the turns generated by student models submitted to the \textsc{strict} track of the BabyLM Challenge. Our motivation is to investigate whether the contingency of multi-turn dialogues can be improved by interaction.

We begin by analysing text generation by previous submissions to the BabyLM Challenge in two settings. First, we analyse the capabilities of models in generating continuations to an input prompt. Models appear to struggle to respond directly to prompts, as shown by a comparison of outputs from BabyLlama \citep{timiryasov-tastet-2023-baby} and our baseline model that we pretrain. This is a 125M parameter Open Pre-Trained Transformer (OPT) \citep{zhang2022opt} trained on 100M words from the \textsc{strict} BabyLM Corpus \citep{charpentier2025babylmturns3papers}.   

\begin{quote}
\textbf{Q:} What do you like to do in summer?\\
\textbf{A: (BabyLlama)} What do you like to do in summer? Well, the winter season is about to begin. The months are longer than usual, and the temperatures are hotter. The months are hotter and colder. So, the months are longer and – \\
\textbf{A (OPT-100M-base):} What's the matter?\\
Nothing.\\
That's all I'm doing.\\
That's what I told you.\\
Look, I'm getting so angry.\\
If you're thinking about your mother...\\
You're gonna get out of...
\end{quote} 

Impressionistically, these LLM outputs show poor contingency with the student model -- failing to directly answer the prompt. We extend our preliminary analysis to investigate multi-turn dialogic interaction betweeen BabyLlama \citep{timiryasov-tastet-2023-baby} and different Teacher LLMs (Gemini 2.5 Flash, Claude 3.7 Sonnet). Dialogues are presented in \textit{Appendix} \ref{app:samples} and show similar behaviour to the examples above. In these preliminary experiments, we see improvements from introducing a \textbf{meta-prompt} which provides instructions to the Teacher Model -- specifying its behaviour as a caregiver, the characteristics of the learner, the goals of the dialogues, and the ideal characteristics of multi-turn interaction.   

%Evaluation of 2024 BabyLMs on Entity Tracking Tasks 
% Qualitative Analysis/Motivation for the outputs of the BabyLMs – even with top-k/temperature 
% Utterance Type (discourse level) – how hard it is to generate dialogue. 
%Entity Annotations 

\subsection{Evaluating BabyLM Coherence and Contingency in Multi-Turn Dialogues}

We go beyond impressionistic evaluation of the contingency of BabyLM outputs as a child/student in multi-turn Teacher-Student Interaction. \citet{agrawal2024automatic} provide an automatic framework for evaluating contingency in CHILDES \citep{macwhinney1985child}. Their evaluation consists of three components: metrics that tag speech-act congruence, semantic alignment between turns (measured using embeddings), and repetition (measured using SpaCy). 

We propose that this evaluation strategy is potentially ill-suited to BabyLMs trained on 100M words, since our preliminary analysis shows that they are able to generate largely grammatical strings but struggle with the essential characteristics of contingency.
%– helping interlocutors generate interventions that connect with the topic of the conversation.  
Contingency, however, is an abstract concept that builds on more concrete linguistic elements, including lexical richness and cohesion. Considering the strong overlap between the notion of contingency and coherence, as defined in Linguistics\footnote{``The state of being logically consistent and connected" \cite{fetzer2012textual}. It depends on a number of factors, including explicit cohesion cues.}, our framework relies on automatic metrics for the analysis of cohesion. 

We introduce two contributions to evaluate contingency in multi-turn dialogues between a Teacher LLM and BabyLM. First, in \textit{Section} \ref{alignment_dataset}, we develop our  \name Alignment Dataset based on discourse cohesion metrics. Secondly, we supplement this with human evaluation following \citet{galvan-sosa-etal-2025-rubriks}'s Rubrik for evaluating LLM-generated text and explanations on the outputs of multi-turn Teacher-Student interaction. 

%adapting Rubrik – language + content dimensions --> some desirable features (basically overlaps with this MT-Bench-101:  https://arxiv.org/pdf/2402.14762).
% clear definition + examples for prompts for COHERENCE and COHESION in a structured few-shot prompt 

\subsection{The \name Alignment Dataset} \label{alignment-dataset}
\label{alignment_dataset}
We capture text complexity differences through five complementary perspectives: semantic ambiguity, discourse connectives, syntactic complexity, cohesion, and lexical complexity. We draw on the \texttt{Switchboard Dialog Act Corpus} to compute different complexity metrics based on these five categories. For a sample, see~\Cref{app:sample_dataset}. We retrieve \textbf{lexical richness} \cite{vajjala2012improving}, type–token ratio (TTR), moving‑average TTR (MATTR), and mean polysemy scores (mPOLY) as proxies for semantic ambiguity. \textbf{Discourse connectives} were quantified by the total number of connectives and the frequency of additive (e.g., ``and’’, ``also’’), adversative (e.g., ``but’’, ``however’’), and causal (e.g., ``because’’, ``therefore’’) subtypes \cite{pitler-nenkova-2009-using}. 

\textbf{Syntactic complexity} was measured by mean sentence length and mean clauses per sentence as indicators of structural elaboration \cite{chen-zechner-2011-computing}. \textbf{Cohesion} was assessed via lexical and grammatical overlap between adjacent sentences (content‑word overlap and verb overlap) and by verb‑tense repetition computed with TAACO‑ and NLTK‑based taggers to capture temporal consistency. \textbf{Semantic and discourse features} included mean age of acquisition (how early words are typically learned), mean CEFR level (Common European Framework of Reference), concept density (distinct concepts per sentence), and an overall narrativity score indexing the extent to which a text exhibits narrative‑like discourse (see~\Cref{app:complexity_metrics}).

We use the 100M word Switchboard Corpus as a large-scale resource for metric estimation. 
This corpus contains transcribed English telephone conversations between speaker pairs in North America \citep{godfrey1992switchboard}.
%In this dataset, speaker turns are concatenated without explicit boundaries, and interlocutors are only implicitly distinguished. To recover the dialogic structure, 
We apply a turn segmentation procedure: utterances are first separated by speaker ID (A or B), consecutive utterances from the same speaker are merged into a single turn, and a ``turn'' is defined as any sequence of text transcribed from one speaker until there is a change of speaker. For dataset annotation, we sample exactly five turns per speaker, truncating the dialogue at that point and continuing segmentation across the whole corpus. Firstly, we compute the complexity metrics across these dialogues to compare metric distributions across speakers (see~\Cref{fig:placeholder}).

\subsection{Manual Evaluation} \label{manual-eval}

The main limitation of the metrics presented in \textit{Section}~\ref{alignment_dataset} is that they were designed primarily for analyzing texts, narratives, and written discourse rather than conversational dialogue.
To address this gap, we adapted the framework proposed by \citet{galvan-sosa-etal-2025-rubriks} for explainability evaluation, which separates assessment into language and content dimensions. 

Language features included \textsc{grammaticality (grm)}, \textsc{word choice (wch)}, and \textsc{cohesion (coh)}, while content features encompassed \textsc{conciseness (cnc)}, \textsc{appropriateness (app)}, and \textsc{coherence (cor)}. 
These feature definitions were adapted from an explainability context to the assessment of conversational contingency.

\section{ \name Methodology}

\subsection{Rewarding Cohesive Response in Multi-Turn Interaction}

Here we investigate two complementary training settings building on preference-based tuning. Experiment 1 uses a fixed teacher to generate improved continuations for student outputs, forming preference triples that fine-tune an OPT-style causal LM \citep{zhang2022opt} with a Reference-Free Preference Objective.  From each Switchboard dialogue in the \name Alignment Dataset (\textit{Section} \ref{alignment_dataset}), we extract one round (two turns) and append the next-speaker prefix to form a continuation prompt; the student samples a reply, and the teacher (\texttt{Llama-3.1-8B-Instruct}) produces a higher-quality alternative under instructions that discourage copying and enforce concise, coherent turns. We filter low-quality teacher outputs with automatic repetition checks, then optimize the student with an odds-ratio style preference objective in five disjoint iterations (dataset slices), carrying weights forward each round. In line with the idea of trial-and-demonstration tuition, this setting treats the teacher's alternative as a soft target that shapes the student's conversational form.

\begin{figure*}[!ht]
    \centering
    \begin{minipage}{0.7\textwidth} 
        \centering
        \includegraphics[width=\textwidth]{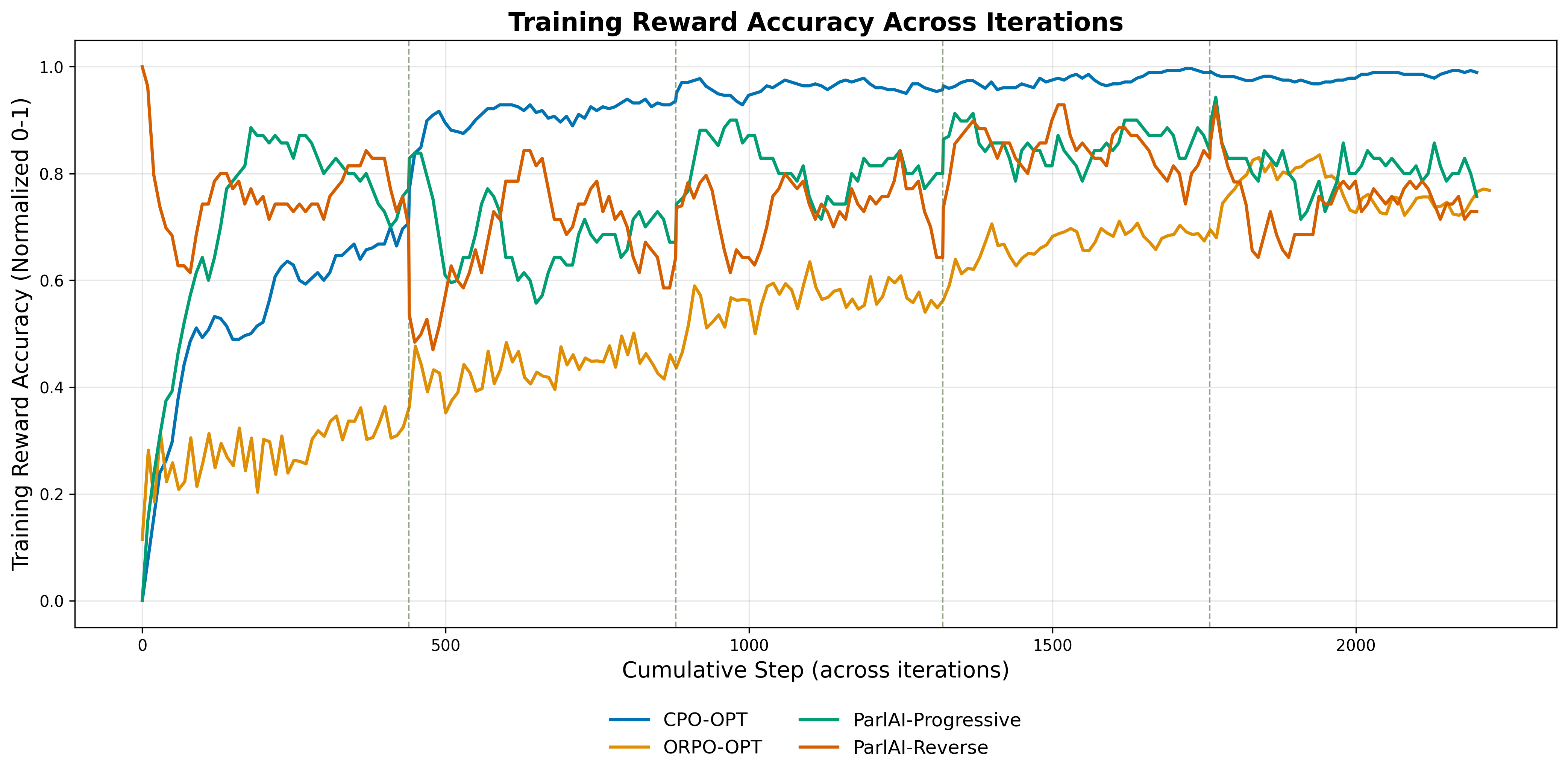}
    \end{minipage}%
    \hfill
    \begin{minipage}{0.28\linewidth} % Caption width
        \caption{Reward accuracy during post-training of OPT (with 1024 sequence length) with CPO/ORPO (Experiment 1) and Progressive/Regressive CEFR (Experiment 2).}
        \label{fig:reward_accuracy_1}
    \end{minipage}
\end{figure*}

\subsection{Adaptively-Decoded Teacher Demonstrations in a BabyLM's Zone of Proximal Development (ZPD)}

While the benefits of including Child Directed Speech (CDS)  in pretraining corpora are contested \citep{feng-etal-2024-child,padovani2025child}, we attempt to lexically constrain the output from a Teacher LLM according to lexical curricula based on the Common European Framework of References for Language (CEFR) \citep{cefr-companion-2020}. We hypothesise that this might simulate more contingent input from a caregiver and thus be a cognitively-inspired mechanism to create \textit{opportune adaptive learning moments} for learners \citep{masek2021language} in a teacher-student interaction. Experiment 2 replaces the teacher with a controllable ParlAI BlenderBot 3B model\footnote{\url{https://huggingface.co/facebook/blenderbot-3B}} and imposes a curriculum over linguistic complexity via CEFR levels. Motivated by the concept of a ZPD -- learning is maximized when input difficulty is just beyond current independent performance -- we constrain the teacher to generate at successive CEFR levels (A2→B1→B2→C1→C2) and, in a separate run, the reverse order, using the CEFR descriptors to operationalize difficulty level.

\section{Experiments}

\subsection{Non-Interactive Baselines}

\subsubsection{Training Datasets}

For our initial experiments we considered a range of pre-training datasets; here, we report experiments for OPT models trained on the \textsc{Strict} BabyLM Corpus. In preliminary work, we pre-trained models on the \textbf{KidLM Corpus} \citep{nayeem-rafiei-2024-kidlm} + \textbf{BabyLM Corpus}: The KidLM Corpus consists of 50.43M words of high-quality genre-diverse child-directed informational content, largely sourced from news articles.  However, we found that models were unable to generate multi-turn dialogues in our post-training experiments. Pre-training corpora more aligned with those used in pre-trained dialogue agents might be more suitable for Teacher-Student Interaction \citep{zhang2020dialogpt}.

\begin{table*}[!ht]
\centering
\scriptsize
\setlength{\tabcolsep}{3pt} % tighten column padding (default ~6pt)
%\resizebox{\textwidth}{!}{%
\begin{tabular}{lccccccccccc}
 \toprule
\textbf{Model} & \textbf{BLIMP} & \textbf{BLIMP-S} & \textbf{COMPS} & \textbf{Entity} & \textbf{EWOK} & \textbf{Eye-Track} & \textbf{Self-Paced} & \textbf{WUG-Adj} & \textbf{WUG-Past} & \textbf{Average} \\
cpo-opt-4096 & 55.59 & 48.08 & 50.27 & \textbf{40.9} & 49.82 & 0.26 & \textbf{0.16} & 0.61 & 0.0 & 27.299 \\
cpo-opt-1024 & 75.74 & 67.73 & 55.51 & 26.69 & 51.36 & 0.33 & 0.03 & 0.66 & 0.04 & 30.899 \\
orpo-opt-1024 & 75.04 & 66.2 & \textbf{56.28} & 26.98 & 51.42 & 0.2 & 0.02 & 0.68 & 0.06 & 30.764 \\
orpo-opt-4096 & 55.71 & 47.73 & 50.13 & 40.75 & 49.33 & 0.2 & 0.1 & 0.65 & -0.02 & 27.176 \\
\midrule
orpo-opt-100M-2048-preprocess & 71.41 & 54.57 & 53.92 & 20.84 & 50.39 & \textbf{1.07} & 0.02 & 0.68 & -0.02 & 28.098 \\
cpo-opt-100M-2048-preprocess & 71.43 & 54.65 & 53.95 & 20.83 & 50.3 & \textbf{1.07} & 0.02 & 0.68 & -0.02 & 28.101 \\
opt-base & 70.45 & 55.18 & 54.28 & 24.1 & 50.83 & 0.45 & 0.03 & \textbf{0.69} & \textbf{0.25} & 28.473 \\
 \midrule
opt-cefr-iteration1 & \textbf{75.96} & 67.73 & 55.9 & 26.78 & \textbf{51.54} & 0.2 & 0.05 & 0.67 & 0.02 & \textbf{30.983} \\
opt-cefr-iteration2 & 75.6 & 67.84 & 55.54 & 26.92 & 51.19 & 0.19 & 0.06 & 0.66 & 0.03 & 30.892 \\
opt-cefr-iteration3 & 75.48 & 67.36 & 55.4 & 27.08 & 51.33 & 0.19 & 0.06 & 0.66 & 0.02 & 30.842 \\
opt-cefr-iteration4 & 75.27 & 67.59 & 55.39 & 26.79 & 51.49 & 0.19 & 0.06 & 0.65 & 0.03 & 30.829 \\
opt-cefr-iteration5 & 75.18 & 67.22 & 55.44 & 26.54 & 51.23 & 0.19 & 0.06 & 0.65 & 0.03 & 30.727 \\
\midrule
%opt-cefr-reverse-iteration1 & 75.91 & 67.73 & 55.9 & 26.73 & 51.49 & 0.2 & 0.05 & 0.67 & 0.02 & 30.967 \\
%opt-cefr-reverse-iteration2 & 75.57 & 67.86 & 55.55 & 26.81 & 51.28 & 0.19 & 0.06 & 0.66 & 0.03 & 30.89 \\
%opt-cefr-reverse-iteration3 & 75.46 & \textbf{67.9} & 55.41 & 26.72 & 51.32 & 0.19 & 0.06 & 0.66 & 0.02 & 30.86 \\
opt-cefr-reverse-iteration4 & 75.32 & 67.7 & 55.43 & 26.51 & 51.5 & 0.19 & 0.06 & 0.65 & 0.02 & 30.82 \\
opt-cefr-reverse-iteration5 & 75.25 & 67.08 & 55.45 & 26.41 & 51.43 & 0.19 & 0.06 & 0.65 & 0.03 & 30.728 \\
\midrule
 \bottomrule
\end{tabular}
%}
\caption{Evaluation results across different BabyLM Evaluation benchmarks (BLiMP, BLiMP Supplement, COMPS, Entity Tracking, EWoK, Eye Tracking and Self-Paced Reading Scores, WUG Adjective Nominalisation and Past Tense) for models in Experiments 1 and 2 compared to baselines. \textbf{Bolded} scores indicate highest accuracy (spearman rho correlation for Wug).}
\label{tab:evaluation-results}
\end{table*}

\begin{table*}[!ht]
\centering
\small
\setlength{\tabcolsep}{3pt} % tighten column padding (default ~6pt)
%\resizebox{\textwidth}{!}{
\begin{tabular}{lcccccc}
\toprule
\bf Model & \bf AoA & \bf CEFR & \bf TTR & \bf Rep. & \bf Overlap & \bf Norm.\ Avg \\

\midrule
cpo-opt-4096 & 4.523 & 1.219 & 0.464 & 0.797 & 0.046 & 0.389 \\
cpo-opt-1024 & 5.011 & 1.408 & \bfseries 0.624 & 0.946 & 0.044 & \bfseries 0.496 \\
orpo-opt-1024 & 4.813 & 1.343 & 0.604 & 0.898 & 0.066 & 0.459 \\
orpo-opt-4096 & 4.487 & 1.228 & 0.473 & 0.879 & 0.075 & 0.346 \\
\midrule
orpo-opt-100M-2048-preprocess & 5.123 & 1.415 & 0.582 & 0.777 & 0.078 & 0.440 \\
cpo-opt-100M-2048-preprocess & 5.087 & 1.373 & 0.620 & 0.851 & 0.068 & 0.436 \\
opt-base & \bfseries 5.214 & \bfseries 1.468 & 0.590 & 0.881 & \bfseries 0.082 & 0.425 \\
\midrule
opt-cefr-iteration1 & 4.813 & 1.325 & 0.627 & 0.803 & 0.044 & 0.386 \\
opt-cefr-iteration2 & 4.802 & 1.371 & 0.610 & 0.870 & 0.053 & 0.454 \\
opt-cefr-iteration3 & 4.782 & 1.305 & 0.606 & 0.904 & 0.082 & 0.469 \\
opt-cefr-iteration4 & 4.951 & 1.328 & 0.599 & 0.887 & 0.062 & 0.465 \\
opt-cefr-iteration5 & 4.819 & 1.328 & 0.572 & 0.909 & 0.056 & 0.465 \\

opt-cefr-reverse-iteration4 & 4.948 & 1.331 & 0.620 & \bfseries 0.942 & 0.062 & 0.458 \\
opt-cefr-reverse-iteration5 & 4.929 & 1.322 & 0.596 & 0.867 & 0.063 & 0.399 \\
\bottomrule
\end{tabular}
%}
\caption{Evaluation Results on Cohesion Metrics of \name Models from Experiment 1 and 2 (CEFR‐based progressive and reverse iterations) against baseline checkpoints. Metrics: AoA (Age of Acquisition), CEFR (mean CEFR level), TTR (type–token ratio), Rep. (verb tense repetition), Overlap (content word overlap), and Norm. Avg (normalized average). Additional evaluation results can be found in \textit{Appendix} \ref{detailed-results-appendix}.}
\label{tab:evaluation-results_v2}
\end{table*}

\subsubsection{Architectures}

We train a 125M OPT architecture with warm-up and a sequence length of \(1024\), which is found by \citet{salhandiehlmartinez} to be an optimal sequence length for pre-training BabyLMs. We also experiment with sequence lengths of \(4096\). See~\Cref{app:model_settings} for detailed experimental settings.

\subsection{Experiment 1: Preference-Free Optimisation }

Contrastive Preference Optimization (CPO) \citep{xu2024contrastive} and Monolithic Odds Ratio Preference Optimization (ORPO) \citep{hong-etal-2024-orpo} are two recent approaches for aligning language models with human preferences, but they differ fundamentally in methodology and applicability. 

CPO extends Direct Preference Optimization (DPO) to train models to avoid producing translations that are adequate but suboptimal, addressing two key limitations of supervised fine-tuning (SFT): the performance ceiling imposed by reference-quality data and the lack of mechanisms to penalize disfavoured outputs. By leveraging contrastive comparisons between preferred and disfavoured outputs, CPO can be applied beyond machine translation to general domains such as dialogue.  Under CPO, the student model would optimize to avoid generating replies that are adequate but suboptimal compared to the teacher’s higher-quality alternative. This implies that the model might produce outputs that are conservatively aligned with the teacher, focusing on minimizing the contrastive loss derived from the teacher’s preferred continuation. As a result, CPO is likely to yield responses that closely match the teacher’s style and content, potentially at the cost of reduced diversity or creativity in dialogue.  In contrast, ORPO eliminates the need for a separate reference model by directly optimizing the odds ratio between favoured and disfavoured outputs within SFT. ORPO's monolithic formulation directly optimizes an odds-ratio preference objective without relying on a separate reference model, integrating the preference signal into the student’s supervised fine-tuning process. Consequently, ORPO can more efficiently incorporate the teacher’s preferred turn while allowing the student greater flexibility in phrasing, leading to responses that retain coherence and conciseness while exploring alternative valid formulations.  We hypothesise that  ORPO may converge faster and produce a wider variety of acceptable replies across the five disjoint iterations, whereas CPO emphasizes stricter adherence to the teacher’s guidance.

\subsection{Experiment 2: Adaptively-Decoded Teacher Model}
For our Teacher Model we follow \citet{tyen-etal-2022-towards} by adaptively decoding the difficulty of messages generated by a BlenderBot 3B model \footnote{\url{https://huggingface.co/facebook/blenderbot-3B}} according to the CEFR language proficiency framework \footnote{\url{https://github.com/WHGTyen/ControllableComplexityChatbot}}. This Controllable Complexity Teacher Model considers multiple candidate messages, before selecting the most appropriate one. We follow the default settings of \citet{tyen-etal-2022-towards} for re-ranking, except we use a smaller beam search size of 5. This generates 5 candidate messages from the Teacher Model for each turn. \citet{tyen-etal-2022-towards} train a regressor to predict the CEFR level of sentences. When the chatbot is in use, the regressor will predict the CEFR level of all candidate messages, allowing us to compute a score that combines the original ranking and the predicted CEFR. This score will then be used to re-rank the candidates, and the top candidate message will be sent to the user. Preference pairs (teacher “chosen” vs.\ student “rejected”) are then used to update the student with a contrastive preference-optimization objective, allowing us to test whether training by complexity --via a CEFR-aware teacher -- better aligns the student's dialogue behaviour with coherent, level-appropriate responses.

\section{Evaluation}

\subsection{Task Evaluation and Post-Training Accuracy}

We evaluate OPT models with sequence lengths of \(\{1024, 4096\}\) on the BabyLM Evaluation Pipeline \citep{charpentier2025babylmturns3papers}. Results are shown in Table \ref{tab:evaluation-results}. The first few rows show evaluation results for Experiment 2 with progressive CEFR alignnent (each iteration is where we progressively increase the CEFR level for adaptive decoding from the Teacher Model). The next few rows compare our OPT models with ORPO \citep{hong-etal-2024-orpo}  and CPO \citep{xu2024contrastive}. We also plot the accuracy of ORPO and CPO on the \name Alignment dataset in \textit{Figure} \ref{fig:reward_accuracy_1}. Additional figures are found in the \textit{Appendix} \ref{training-metrics}. 

\subsection{Text Generation Evaluation}

We evaluate our models using Cohesion Metrics and different meta-prompts that aim to simulate differences in dialogue generation characteristics. 

\textbf{Cohesion Metrics.} Based on Table \ref{tab:evaluation-results_v2}, cpo-opt-1024 achieves the highest normalized average score (0.496) among the CPO/ORPO variants, with strong lexical diversity (TTR = 0.624) and high repetition control (Rep. = 0.946), indicating robust overall performance. There are inconsistent benefits of CEFR-alignment. Worse performance might be due to limited beam search since \citet{tyen-etal-2022-towards} generate 20 responses per turn.  

\textbf{Human Evaluation.}  Following \citet{galvan-sosa-etal-2025-rubriks}'s approach, each feature was manually assessed in a binary manner (yes/no) for each dialogue in the evaluation set generated using 5 conversation starters that were consistent with our preliminary dialogue generation.

\begin{table}[t!]
\centering
\scriptsize
\begin{tabular}{p{1.8cm}>{\centering\arraybackslash}p{0.4cm}>{\centering\arraybackslash}p{0.5cm}>{\centering\arraybackslash}p{0.5cm}>{\centering\arraybackslash}p{0.5cm}>{\centering\arraybackslash}p{0.5cm}>{\centering\arraybackslash}p{0.5cm}}
\toprule
{\bf Teacher} & {\bf GRM}& {\bf WCH}& {\bf COH}& {\bf CNC}& {\bf APP}& {\bf COR}\\\midrule
Dialogue 1 & \ding{51} & \ding{55} & \ding{51}& \ding{55} & \ding{55} & \ding{55}\\
Dialogue 2 & \ding{55} & \ding{55} & \ding{55}& \ding{55} & \ding{55} & \ding{55}\\
Dialogue 3 & \ding{51} & \ding{55} & \ding{51} & \ding{55}& \ding{55} & \ding{55}\\
Dialogue 4 & \ding{51} & \ding{55} & \ding{51}& \ding{55} & \ding{55} & \ding{55}\\ \midrule
{\bf Student} & {\bf GRM}& {\bf WCH}& {\bf COH}& {\bf CNC}& {\bf APP}& {\bf COR}\\ \midrule
cpo-opt-1024 & \ding{51} & \ding{55} & \ding{51}& \ding{55} & \ding{55} & \ding{55}\\
cpo-opt-4096 & \ding{55} & \ding{55} & \ding{55}& \ding{55} & \ding{55} & \ding{55}\\
orpo-opt-1024 & \ding{51} & \ding{55} & \ding{51}& \ding{55} & \ding{55} & \ding{55}\\
orpo-opt-4096 & \ding{55} & \ding{55} & \ding{55}& \ding{55} & \ding{55} & \ding{55}\\
\bottomrule 
\end{tabular}
\caption{ Qualitative judgements of Grammaticality, Word choice, Cohesion, Conciseness, Appropriateness and Coherence. Dialogues 1 - 5 refer to LLama 3.1B with corresponding student model.}
 \label{manual_eval}
\end{table}

While most of the dialogues were judged to be grammatical and cohesive, they failed to meet the rest of the features of contingency. Table \ref{manual_eval} reports binary human judgements on 10 multi-turn dialogues with 8 turns generated between Llama-3.1B Teacher and four student models (\texttt{cpo-1024, cpo-4096, orpo-1024, orpo-4096}). This highlights the inherent complexity of conversational text, where lexical overlap within individual turns does not necessarily indicate that the dialogue as a whole achieved contingent interaction. Here, \texttt{cpo-opt-1024} achieves the best overall performance. 

\textbf{Meta-Prompts.} Table \ref{tab:age-segment-metrics} in the \textit{Appendix} shows generation across interactions between teacher and student generated by \texttt{cpo-opt-1024} with meta-prompts to the Teacher Model to generate dialogue starters based on specified age roles of the student model. Across most metrics, the 3–4 years group exhibits a more complicated linguistic profile in comparison to 2–3 years, with the highest Age of Acquisition (AoA) and CEFR level. Younger groups (6–11 months and 18–23 months) often occupy a middle ground, but show more overlap in aggregated data. With increasing interaction, we observe a drop in TTR and lexical richness (AoA/CEFR), and increases in cohesion and repetition (i.e., Overlap/Rep.).

\section{Discussion}

Our preliminary analysis of outputs from \textsc{Strict} BabyLMs highlights a persistent gap between grammaticality and the communicative capabilities of BabyLMs, and we have presented contingency as one way to potentially improve interactional abilities.
%: the ability to produce contextually appropriate and temporally coherent responses in multi-turn exchanges. 
Our different experimental setups explore how \textbf{teacher demonstrations} can be utilised in multi-turn interaction.  Experiment 1 uses an interactive setup that provides a measurable quantitative and qualitative improvement in turn-level coherence, lexical continuity, and grammatical repair across multi-turn Teacher-Student Interactions. It is possible to distinctly interpret ORPO and CPO post-training pipelines used in \name in Vygotskian terms – both define and regulate a dynamic “scaffolded” learning region for the BabyLM (a Zone of Proximal Development) where communicative competence can be acquired through guided interaction.  

The interplay of reward signals and policy constraints determines how far the student BabyLM may deviate from the behaviour of the Teacher LLM, while still being reinforced for progress toward more contingent, coherent, and human-like dialogue generation. The noticeable gains of the CPO reward model compared to ORPO are significant. CPO constrains the policy update to remain close to teacher demonstrations, effectively keeping the BabyLM’s learning trajectory within a tightly scaffolded region of its ZPD. Throughout post-training, CPO anchors the updates of the BabyLMs more strongly than ORPO, potentially preventing premature drift into ungrounded or incoherent communicative behaviours. ORPO encourages exploration along preference gradients that are partially decoupled from the teacher’s demonstrations, which could promote long-term generalisation and independence but also increases the likelihood of divergence from high-quality exemplars early on, leading to noisier learning dynamics or inconsistent contingent behaviour. In developmental terms, this potentially suggests that \textbf{BabyLMs might benefit from strong scaffolding via feedback that rewards improvement and maintains the student model’s proximity to Teacher performance}. 

In contrast,  Experiment~2 revealed limited performance gains when the BabyLM is trained solely on static lexically-constrained Teacher Demonstrations without ongoing preference feedback. Although the model maintained grammatical competence and modestly improved surface-level coherence, it showed little advancement in deeper measures of contingency, such as pragmatic relevance and discourse-level alignment. This asymmetry suggests a crucial distinction between demonstrative and interactive scaffolding: while demonstrations expose the learner to appropriate communicative forms, they do not convey the adaptive feedback necessary to internalise when and why these forms should be used. Without the dynamic reinforcement provided in Experiment~1, the BabyLM might remain confined within its ZPD; capable of imitation, but unable to generalise beyond it. Further controlled experimentation is needed to confirm this hypothesis: for example, investigating different types of adaptive feedback that can improve contingency.

\section{Conclusion}
Our work demonstrates that contingency -- prompt, direct, and meaningful exchanges -- can be effectively benchmarked and improved in BabyLMs using the \name Teacher–Student framework. Post-training with a carefully designed alignment dataset leads to more grammatical and cohesive multi-turn responses, while adaptive teacher decoding offers limited additional gains. The conditions for contingent dialogues from a BabyLM improve with \textbf{interactive scaffolding} and \textbf{adaptive feedback}, highlighting the benefits of continued ongoing, context-sensitive guidance that aligns learning signals with clear communicative goals. These results underscore the value of targeted post-training for enhancing dialogue quality and establish contingency as a meaningful and challenging objective for future BabyLM research.

\section*{Limitations}
While \name introduces a cognitively-motivated framework for enhancing contingency in small language models, several limitations constrain the generality and interpretability of our findings.

\textbf{Post-Training Data and Domain.} Our alignment dataset is derived exclusively from the Switchboard Dialog Act Corpus \citep{stolcke2000dialogue}, which, although large and richly annotated, represents a narrow sociolinguistic domain—adult telephone conversations in American English. Consequently, the patterns of contingency learned during post-training may not generalise to other interactional contexts such as narrative discourse, spontaneous child-directed speech, or multilingual dialogue. Future work should extend our approach to corpora that more closely resemble early caregiver–child interactions or include non-Western varieties of English.

\textbf{Limited Interpretability of Post-Training} Reward-based fine-tuning may conflate linguistic and stylistic signals, making it challenging to disentangle which aspects of contingency are actually learned.

\textbf{Experiments only with one Teacher Model} All Student models were trained with feedback from a single Teacher LLM (Llama-3.1-8B-Instruct). This limits the  robustness of our claims about the resulting contingent behaviour, as improvements may reflect stylistic imitation or alignment to that specific model’s discourse patterns rather than generalized contingent competence. Investigation with more Teacher Models ecologically valid estimation of the Student’s Zone of Proximal Development (ZPD).

\textbf{Combination of Automatic and Human Evaluation} Cohesion-based metrics (e.g., lexical overlap, verb repetition, CEFR-based lexical complexity) were originally developed for written text and do not fully capture pragmatic or conversational aspects of contingency such as repair, implicature, or turn-taking latency. Although we supplement these with human evaluation and dedicated significant effort to selecting automated metrics for evaluating dialogues, the resulting measures are imperfect proxies for the dynamic adaptivity that characterises natural dialogue.

\section*{Acknowledgements}
The ALTA Institute authors are supported by Cambridge University Press \& Assessment. Donya Rooein’s research is supported through the European Research Council (ERC) under the European Union’s Horizon 2020 research and innovation program (No. 949944, INTEGRATOR). We would like to extend our thanks to Ethan Wilcox and Leshem Choshen, in particular, for their support during the review process, alongside the other BabyLM Workshop Organisers.  This research was performed using resources provided by the Cambridge Service for Data Driven Discovery (CSD3) operated by the University of Cambridge Research Computing Service, provided by Dell EMC and Intel using Tier-2 funding from the Engineering and Physical Sciences Research Council (capital grant EP/T022159/1), and DiRAC funding from the Science and Technology Facilities Council.

%\textbf{GPT-2 Baseline \citep{goriely-etal-2024-babble}:} \citet{goriely-etal-2024-babble} achieve a strong performance on the 2024 BabyLM Leaderboard with a GPT2 base architecture with Byte-Pair Encoding with no word boundaries with a dedicated “utterance boundary” token \texttt{UTT\_BOUNDARY}  at the start of each sentence. This implicitly serves as an end-of-sentence token when sentences are collated. Whitespace is removed during pre-processing to allow the model to learn “super-words” that cross token boundaries. We use a BPE tokeniser with a 16K vocabulary. \textbf{GPT-BERT \citep{charpentier-samuel-2024-bert}:} The GPT-BERT architecture attempts ot 

\bibliography{custom}
\bibliographystyle{acl_natbib}

\appendix
\onecolumn

\section{Sample Dialogues}
\label{app:samples}

%\subsection{Preliminary Analysis}
\subsubsection*{Gemini 2.5 Flash-BabyLLama}
\begin{quote}
\textbf{Q:} What do you like to do in summer?\\
- It was a good day.\\
- This is a new day.\\
- What is it, a big party?\\
- It's just, like, a real big thing.\\
- What's this?\\
- It's really
\end{quote}

\subsubsection*{Claude 3.7 Sonnet-BabyLLama}
\begin{quote}
\textbf{Q:} What do you like to do in summer?\\
- It was a good day.\\
- This is a new day.\\
- What is it, a big party?\\
- It's just, like, a real big thing.\\
- What's this?\\
- It's really
\end{quote}

\newpage 
\section*{Caregiver–Kid Communication Dialogue Meta-Prompt} \label{meta-prompt}

%\subsubsection*{ Meta-Prompts}

\begin{quote}
\textbf{Instructions:} Assume the role of a caregiver. I will assume the role of a kid that's \texttt{<INSERT AGE>} years old. Your goal is to test the kid's communication skills through a dialogue, where you need to take the lead. Whenever it is your turn, consider that your responses should be no longer than one sentence.  

You should aim to end the dialogue within 10 turns total (5 turns you, 5 turns me). When we reach the end, print a summary of our dialogue.  

\textbf{Start by asking:} What do you like about summer?
\end{quote}

\newpage 
\section{Sample \name Alignment Dataset}
\label{app:sample_dataset}

\lstdefinelanguage{json}{
    basicstyle=\ttfamily\small,
    numbers=left,
    numberstyle=\tiny\color{gray},
    stepnumber=1,
    numbersep=5pt,
    showstringspaces=false,
    breaklines=true,
    % frame=single,        % <-- removed
    backgroundcolor=\color{gray!5},
    literate=
     *{0}{{{\color{blue}0}}}{1}
      {1}{{{\color{blue}1}}}{1}
      {2}{{{\color{blue}2}}}{1}
      {3}{{{\color{blue}3}}}{1}
      {4}{{{\color{blue}4}}}{1}
      {5}{{{\color{blue}5}}}{1}
      {6}{{{\color{blue}6}}}{1}
      {7}{{{\color{blue}7}}}{1}
      {8}{{{\color{blue}8}}}{1}
      {9}{{{\color{blue}9}}}{1}
      {:}{{{\color{red}{:}}}}{1}
      {,}{{{\color{red}{,}}}}{1}
      {"}{{{\color{black}{"}}}}{1}
}
\begin{figure}[!ht]
\centering
\begin{minipage}{0.95\linewidth}
\begin{lstlisting}[language=json]
{
  "dialogID": "dialog_02341.txt",
  "turns": [
    {
      "speaker": "B",
      "utterance": "And nothing is being done about it. Uh, the laws exist and are frequently upheld in, in, uh, in Appeals Court just because of technicalities and because of maybe small little holes that their defending attorney can find. And it's, it's really getting out of hand in many states."
    },
    {
      "speaker": "A",
      "utterance": "Well, the term technicality. The law enforcement community, uh, uh, you know, has to, has to separate the difference between somebody who is being set up in which, uh, grievous acts are done to, uh, to, you know, to get somebody into a, a situation where they're going to be guilty of, of a crime ..."
    },
    {
      "speaker": "B",
      "utterance": "Well, it seems like well it, it seems as if in the past typically there have been a lot of cases of people being wrongly tried or wrongly punished ...
    },
    {
      "speaker": "A",
      "utterance": "Uh-huh."
    },
    {
      "speaker": "B",
      "utterance": "And where his, old evidence was there, the witnesses were there, the, everything was conclusively pointing to this individual yet"
    }
  ],
}
\end{lstlisting}
\end{minipage}
\caption{Sample of the \name \textsc{Alignment} Dataset}
\label{fig:dialog_json_a}
\end{figure}

\newpage 

\begin{figure}[!ht]
\centering
\begin{minipage}{0.95\linewidth}
\begin{lstlisting}[language=json]

  "meta": {
    "length": 593,
    "ttr": {
      "noun": 0.162852,
      "verb": 0.154903,
      "adj": 0.182672
    },
    "type_token_ratios": [
      {
        "noun_ttr": 0.71,
        "verb_ttr": 0.475,
        "adj_ttr": 0.8571428571428571,
        "lemma_ttr": 0.332794830371567,
        "bigram_lemma_ttr": 0.8155339805825242,
        "trigram_lemma_ttr": 0.9708265802269044,
        "adjacent_overlap_all_sent": 0.1912442396313364,
        "lda_1_all_sent": 0.8396384935744969,
        "repeated_content_lemmas": 0.2116316639741518,
        "repeated_content_and_pronoun_lemmas": 0.2762520193861066
      },
      {
        "noun_ttr": 0.8166666666666667,
        "verb_ttr": 0.4561403508771929,
        "adj_ttr": 0.9444444444444444,
        "lemma_ttr": 0.3525179856115107,
        "bigram_lemma_ttr": 0.8269230769230769,
        "trigram_lemma_ttr": 0.9662650602409638,
        "adjacent_overlap_all_sent": 0.2067796610169491,
        "lda_1_all_sent": 0.8670341452328432,
        "repeated_content_lemmas": 0.1750599520383693,
        "repeated_content_and_pronoun_lemmas": 0.237410071942446
      },
      {
        "noun_ttr": 0.8857142857142857,
        "verb_ttr": 0.782608695652174,
        "adj_ttr": 1.0,
        "lemma_ttr": 0.5279187817258884,
        "bigram_lemma_ttr": 0.9183673469387756,
        "trigram_lemma_ttr": 0.9948717948717948,
        "adjacent_overlap_all_sent": 0.1742424242424242,
        "lda_1_all_sent": 0.8547770311665861,
        "repeated_content_lemmas": 0.116751269035533,
        "repeated_content_and_pronoun_lemmas": 0.182741116751269
      }
    ],
    "sentiment_scores": {
      "polarity": -0.473618,
      "subjectivity": -0.009093,
      "toxicity": 0.189254
    }
  }
}
\end{lstlisting}
\end{minipage}
\caption{Sample of the \name \textsc{Alignment} Dataset}
\label{fig:dialog_json_b}
\end{figure}

\newpage 

\section{Experimental Settings}
\label{app:model_settings}

\subsection{Decoder Settings for Text Generation}
\begin{table*}[!ht]
\centering
\small
\caption{Decoding settings for Student and Teacher Generation.}
\label{tab:appendix-decoding}
\begin{tabular*}{\textwidth}{l l l}
\toprule
\textbf{Component} & \textbf{Parameter} & \textbf{Value} \\
\midrule
Student (child) & \texttt{max\_new\_tokens} & 100 \\
 & \texttt{do\_sample} & True \\
 & \texttt{top\_k} & 50 \\
 & \texttt{top\_p} & 0.95 \\
 & \texttt{temperature} & 0.8 \\
 & \texttt{num\_return\_sequences} & 1 \\
\midrule
Teacher (LLM) & \texttt{max\_new\_tokens} & 50 \\
 & \texttt{do\_sample} & False \\
% (temperature not used when deterministic)
\bottomrule
\end{tabular*}
\end{table*}

\subsection{Training Hyperparameters shared across Experiments}
% =========================
% Training hyperparameters (shared by ORPO and CPO)
% =========================
\begin{table*}[!ht]
\centering
\small
\caption{Preference optimization hyperparameters (ORPO and CPO; identical across experiments).}
\label{tab:appendix-train}
\begingroup
\setlength{\tabcolsep}{3pt}           % default ~6pt; tighten to fit
\renewcommand{\arraystretch}{1.05}    % small breathing room
\begin{tabular*}{\textwidth}{@{\extracolsep{\fill}} l c c c c c c c c}
\toprule
\textbf{per-dev bsz} & \textbf{grad accum} & \textbf{eff.\ bsz} & \textbf{lr} &
\textbf{epochs} & \textbf{warmup} & \textbf{max grad norm} & \textbf{fp16} & \textbf{grad ckpt} \\
\midrule
1 & 8 & 8 & $1\times10^{-6}$ & 1 & 10 & 0.5 & False & True \\
\midrule
\textbf{optimizer} &
\shortstack{\textbf{remove}\\\textbf{unused}\\\textbf{cols}} &
\shortstack{\textbf{drop}\\\textbf{last}} &
\shortstack{\textbf{num}\\\textbf{workers}} &
\shortstack{\textbf{save}\\\textbf{steps}} &
\shortstack{\textbf{eval}\\\textbf{steps}} &
\shortstack{\textbf{logging}\\\textbf{steps}} & & \\
\midrule
\texttt{adamw\_torch} & False & True & 0 & 500 & 500 & 10 & & \\
\bottomrule
\end{tabular*}
\caption{Trainer setup for CPO and ORPO}
\endgroup
\end{table*}

\subsection{ParlAI teacher (CEFR-controlled) configuration}

% =========================
% ParlAI teacher (CEFR-controlled) configuration
% =========================

\begin{table*}[h!]
\centering
\caption{Key hyperparameters for ParlAI ControllableBlender teacher agent.}
\begin{tabular}{ll}
\hline
\textbf{Parameter} & \textbf{Value / Description} \\
\hline
Model Zoo & \texttt{blender\_3B} (BlenderBot 3B) \\
Beam Size & 20 \\
Top-K Sampling & 40 \\
Rerank CEFR Level & dynamically set per ORPO phase (A2/B2/C1) \\
Rerank Tokenizer & \texttt{distilroberta-base} \\
Rerank Model & \texttt{complexity\_model} \\
Rerank Model Device & \texttt{cuda} \\
Inference Mode & \texttt{rerank} \\
Filter Path & \texttt{data/filter.txt} (default) \\
Child Generation Args & max\_new\_tokens=50, do\_sample=True, top\_k=50 \\
 &  top\_p=0.95, temperature=0.8 \\

Teacher Generation Args & max\_new\_tokens=50, do\_sample=False, temperature=0.3 \\
Number of Prompts & 8 (sampled per ORPO iteration) \\
Max Input Length & 512 tokens (child fine-tuning) \\
\hline
\end{tabular}
\label{tab:parlai-hparams}
\end{table*}
\newpage

\subsection{Pretraining Hyperparameters}
\begin{table}[!ht]
\centering
\begin{tabular}{lcc}
\toprule
\textbf{Parameter} & \textbf{Mamba} & \textbf{OPT} \\
\midrule
\texttt{vocab\_size} & 50280 & 50272 \\
\texttt{hidden\_size} & 768 & 768 \\
\texttt{num\_hidden\_layers} & 32 & 12 \\
\texttt{state\_size} & 16 & -- \\
\texttt{expand / ffn\_dim } &  2 & 3072 \\
\texttt{num\_attention\_heads} & -- & 12 \\
\texttt{hidden\_act} & silu & relu \\
\bottomrule
\end{tabular}
\caption{Key default hyperparameters for \texttt{MambaConfig} and \texttt{OPTConfig} as implemented in Hugging Face Transformers.}
\label{tab:key-defaults}
\end{table}

\clearpage 

\section{Linguistic Complexity Metrics} \label{app:complexity_metrics}
\begin{table*}[!ht]
\centering
\small
\label{tab:metrics}
\begin{tabularx}{\textwidth}{@{} l l l X l @{}}
\toprule
\textbf{Metric} & \textbf{Abbrev.} & \textbf{Category} & \textbf{Description} & \textbf{Source} \\
\midrule
Type--Token Ratio  & TTR & Lexical richness & Ratio of unique types to total tokens; indexes vocabulary diversity. & \texttt{TAACO} \\
Moving-Average TTR  & MATTR & Lexical richness & Mean TTR over a sliding window to reduce text-length sensitivity. & \texttt{TAACO}\\
Mean polysemy & mPOLY & Lexical richness & Average meaningfulness scores for words in text & \texttt{CRAT} \\
Total discourse connectives & TDC & Discourse connectives & Count of connective tokens that explicitly link ideas across clauses/sentences. &  \texttt{manual}\\
Additive connectives frequency & ACF & Discourse connectives & Rate of additive connectives (``and'', ``also'', etc.). &  \texttt{manual}\\
Adversative connectives frequency & AdCF & Discourse connectives & Rate of adversative connectives (``but'', ``however'', etc.). & \texttt{Spacy}\\
Causal connectives frequency & CaCF & Discourse connectives & Rate of causal connectives (``because'', ``therefore'', etc.). & \texttt{manual} \\
Mean sentence length & MSL & Syntactic complexity & Average number of tokens per sentence. & \texttt{Spacy} \\
Mean clauses per sentence & MCPS & Syntactic complexity & Average number of clauses per sentence. &  \texttt{Spacy}\\
Content-word overlap (adjacent) & CWO-Adj & Cohesion & Proportion of content lemmas shared between adjacent sentences. &  \texttt{TAACO}\\
Verb overlap (adjacent) & VO-Adj & Cohesion & Verb overlap between adjacent sentences &  \texttt{TAACO}\\
Verb tense repetition (Repetition) & VTR & Cohesion & Share of adjacent sentences with matching verb tense (temporal consistency). &  \texttt{NLTK} \\
% Referential Cohesion Score (Ref. Cohesion) & --- & Cohesion & Composite referential cohesion score & \texttt{manual}\\
Mean age of acquisition  & AoA & Semantic & Average age at which words in the text are typically acquired. & \texttt{CRAT} \\
Mean CEFR level  & CEFR & Semantic & Average CEFR level of words in text. & \texttt{CRAT} \\
Mean familiarity & MFam & Semantic & Average familiarity scores for words in text. & \texttt{CRAT} \\
Concept density & CD & Semantic & Number of concepts per sentence &  \texttt{spaCy}  \\
Narrativity score & Narr & Semantic & Composite narrativity score based on multiple metrics & \texttt{manual} \\
\bottomrule
\end{tabularx}
\caption{Metrics used to assess linguistic complexity of the texts across five different categories.}
\end{table*}

\newpage

Cohesion is computed as the average of all normalised TAACO metrics. Since all TAACO metrics range between 0–1, a simple mean provides an overall score; however, because some metrics consistently score near the top, examining their variance and distribution helps refine weighting. The figure shows the distributions of the 10 selected TAACO metrics.
\begin{figure*}[h!]
    \centering
    \includegraphics[width=\textwidth]{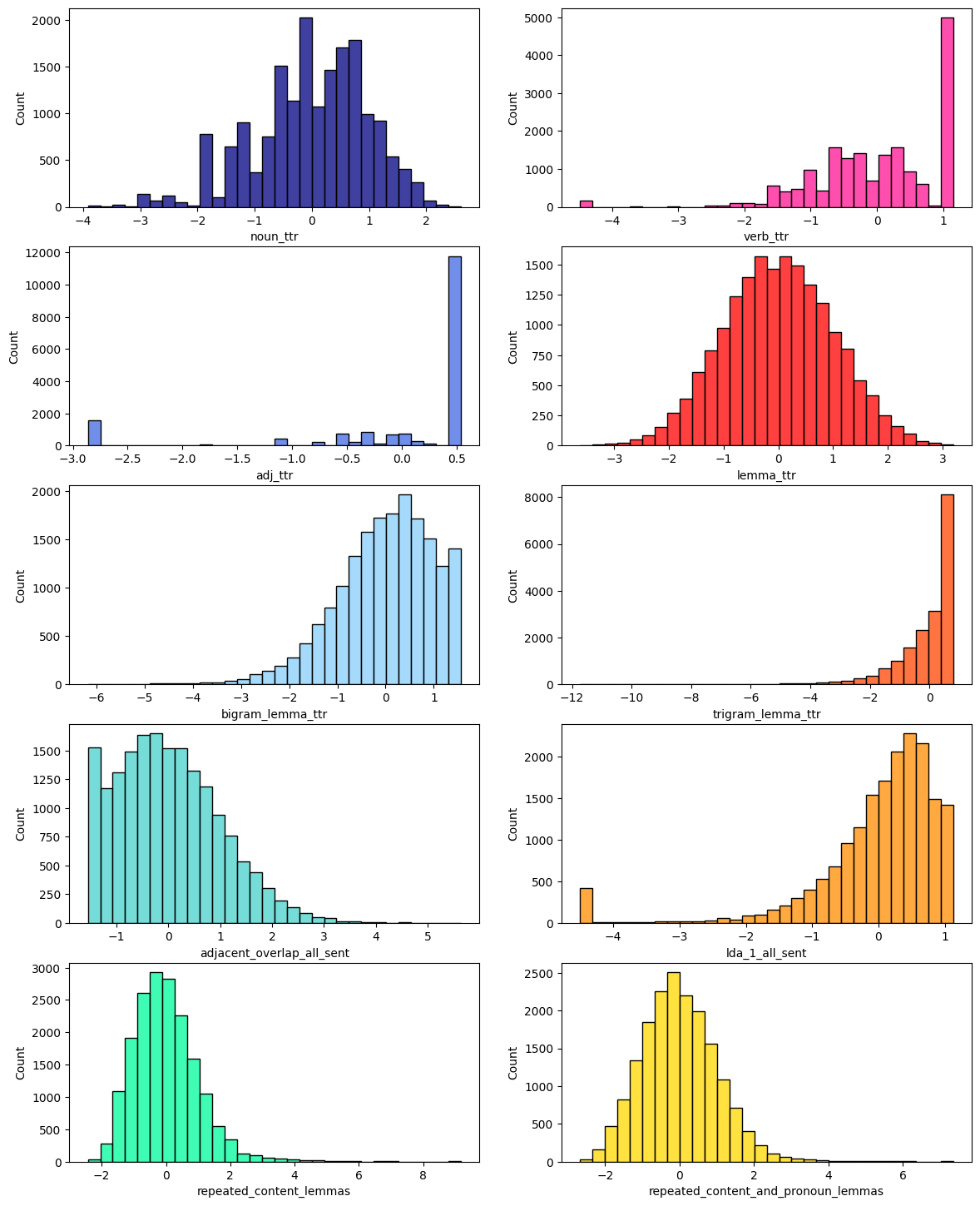}
    \caption{Distributions of the 10 selected TAACO metrics over the Switchboard dataset.}
    \label{fig:placeholder}
\end{figure*}

%\begin{landscape}
\newpage 
\section{Detailed Analysis of Teacher-Student Multi-Turn Dialogues} \label{detailed-results-appendix}

We include a detailed analysis of generated Teacher-Student Multi-Turn dialogues across different lengths.  We include a more detailed breakdown of results summarised in  \textit{Table} \ref{tab:evaluation-results_v2}.  The teacher model, \verb|meta-llama/Llama-3.2-3B-Instruct|, provides guidance and responses, while the student model, (e.g., \verb|babylm-seqlen/opt-1024-warmup-v2|), is prompted and evaluated using the \verb|facebook/opt-125m| tokenizer. We report results with \(\{2,4,6,8\}\) back-and-forth turns, with a maximum of \(\{50,100, 150, 200, 250\}\) tokens per turn.

Our generation scripts include utilities for cleaning generated responses, removing role tokens, unwanted punctuation, and other extraneous symbols, while also identifying banned tokens to avoid during generation. Teacher and student responses are generated using controlled sampling parameters such as top-p, top-k, temperature, and repetition penalties, with the student generation including multiple retry attempts to ensure meaningful output. The main generation function orchestrates multi-turn conversations, alternating between student and teacher turns, starting from a randomly selected conversation starter. The generated dialogues are structured with metadata, turn indices, and clean transcript text, and are finally saved as JSON files in a specified output directory. The script also includes a command-line interface allowing users to specify model IDs, tokenizers, number of turns, maximum token lengths, random seeds, devices, and output paths, making it versatile for experimentation and reproducible dialogue generation.

\subsection{Conversation Starters}

We provide the following conversation starters to generate dialogues between our Student and Teacher models. We apply automated metrics to these models. 

\begin{figure}[!ht]
\centering
\begin{minipage}{0.95\linewidth}
\begin{lstlisting}[language=json]
{
  "STARTERS": [
    "Have you been on any trips recently? Where did you go, "
    "and did anything interesting happen there?",

    "What kind of music do you usually listen to? Do you have "
    "a favorite artist or concert experience you remember?",

    "Do you enjoy cooking at home? What's the best meal you've "
    "made recently, or do you prefer eating out?",

    "Do you have any pets? How long have you had them, and "
    "what do you like most about them?",

    "Do you play any sports or keep active? Have you joined any "
    "teams or tried something new lately?",

    "What's the weather usually like where you live? Does it affect "
    "your plans or the way you spend your weekends?",

    "Have you watched any shows or movies recently? Did you enjoy "
    "them, and would you recommend them to others?",

    "How's work going these days? Have you faced any interesting "
    "challenges or had any funny moments?",

    "Do you have any hobbies you like to spend time on? How did "
    "you get into them, and what keeps you interested?",

    "Do you celebrate any holidays with your family? Are there "
    "any special traditions or funny stories from past celebrations?"
  ]
}
\end{lstlisting}
\caption{Conversation starters used as initial prompts for multi-turn dialogue generation. Each starter is an open-ended question designed to elicit rich responses.}
\label{fig:conversation_starters}
\end{minipage}
\end{figure}

\newpage

\subsection{Extended \Cref{tab:evaluation-results_v2} Results} \label{app:extended}

\begin{table*}[!ht]
\centering
\tiny
\setlength{\tabcolsep}{4pt}
\begin{tabular}{lrrrrrrrr}
\toprule
\textbf{Model} & \textbf{Turns} & \textbf{AoA} & \textbf{CEFR} & \textbf{Overlap} & \textbf{TTR} & \textbf{Rep.} & \textbf{NumCon} & \textbf{NormAvg} \\
\midrule
cpo\_opt\_100M\_2048\_preprocess & 4 & 5.087 & 1.373 & 0.068 & 0.620 & 0.851 & 13.300 & 0.503 \\
cpo\_opt\_base & 4 & 5.214 & 1.468 & 0.074 & 0.590 & 0.881 & 13.100 & 0.556 \\
cpo\_opt\_cosmos & 4 & 5.067 & 1.383 & 0.052 & 0.638 & 0.912 & 14.200 & 0.530 \\
cpo\_opt\_seqlen\_1024\_final\_checkpoint & 4 & 5.011 & 1.408 & 0.044 & 0.624 & 0.946 & 15.600 & 0.536 \\
cpo\_opt\_seqlen\_1024\_progressive\_cefr\_parlai\_iteration1 & 4 & 4.813 & 1.325 & 0.044 & 0.627 & 0.803 & 16.100 & 0.402 \\
cpo\_opt\_seqlen\_1024\_progressive\_cefr\_parlai\_iteration2 & 4 & 4.802 & 1.371 & 0.053 & 0.610 & 0.870 & 16.500 & 0.460 \\
cpo\_opt\_seqlen\_1024\_progressive\_cefr\_parlai\_iteration3 & 4 & 4.782 & 1.305 & 0.082 & 0.606 & 0.904 & 13.600 & 0.497 \\
cpo\_opt\_seqlen\_1024\_progressive\_cefr\_parlai\_iteration4 & 4 & 4.951 & 1.328 & 0.062 & 0.599 & 0.887 & 13.400 & 0.485 \\
cpo\_opt\_seqlen\_1024\_progressive\_cefr\_parlai\_iteration5 & 4 & 4.819 & 1.328 & 0.056 & 0.572 & 0.909 & 13.800 & 0.467 \\
cpo\_opt\_seqlen\_1024\_progressive\_cefr\_reverse\_parlai\_iteration4 & 4 & 4.948 & 1.331 & 0.062 & 0.620 & 0.942 & 14.400 & 0.529 \\
cpo\_opt\_seqlen\_1024\_progressive\_cefr\_reverse\_parlai\_iteration5 & 4 & 4.929 & 1.322 & 0.063 & 0.596 & 0.867 & 15.000 & 0.469 \\
cpo\_opt\_seqlen\_4096\_final\_checkpoint & 4 & 4.523 & 1.219 & 0.046 & 0.464 & 0.797 & 18.100 & 0.286 \\
mamba-sam-seqlen-2048-original & 4 & 4.770 & 1.353 & 0.048 & 0.609 & 0.945 & 15.700 & 0.494 \\
opt-sam-orpo-mamba-2048-step448 & 4 & 4.887 & 1.376 & 0.072 & 0.617 & 0.918 & 16.800 & 0.529 \\
opt-sam-orpo-seqlen-2048-step559 & 4 & 4.948 & 1.365 & 0.052 & 0.615 & 0.866 & 15.100 & 0.474 \\
opt-sam-seqlen-2048-original & 4 & 4.843 & 1.318 & 0.052 & 0.618 & 0.942 & 17.000 & 0.503 \\
orpo\_opt\_100M\_2048\_preprocess & 4 & 5.123 & 1.415 & 0.078 & 0.582 & 0.777 & 14.300 & 0.467 \\
orpo\_opt\_cosmos & 4 & 5.200 & 1.459 & 0.067 & 0.624 & 0.822 & 10.300 & 0.514 \\
orpo\_opt\_seqlen\_1024\_final\_checkpoint & 4 & 4.813 & 1.343 & 0.066 & 0.604 & 0.898 & 17.900 & 0.488 \\
orpo\_opt\_seqlen\_4096\_final\_checkpoint & 4 & 4.487 & 1.228 & 0.075 & 0.473 & 0.879 & 11.500 & 0.371 \\
babylm-seqlen-opt-1024-warmup-v2 & 4 & 4.864 & 1.343 & 0.057 & 0.594 & 0.925 & 16.500 & 0.496 \\
babylm-seqlen-opt-4096-warmup-v2 & 4 & 4.487 & 1.172 & 0.030 & 0.449 & 0.867 & 17.100 & 0.291 \\
cpo\_opt\_100M\_2048\_preprocess & 6 & 5.071 & 1.423 & 0.065 & 0.564 & 0.830 & 19.000 & 0.479 \\
cpo\_opt\_base & 6 & 5.022 & 1.355 & 0.057 & 0.549 & 0.891 & 18.400 & 0.483 \\
cpo\_opt\_cosmos & 6 & 5.123 & 1.457 & 0.049 & 0.560 & 0.912 & 20.300 & 0.526 \\
cpo\_opt\_seqlen\_1024\_final\_checkpoint & 6 & 4.947 & 1.405 & 0.050 & 0.563 & 0.911 & 25.900 & 0.498 \\
cpo\_opt\_seqlen\_1024\_progressive\_cefr\_parlai\_iteration1 & 6 & 4.898 & 1.350 & 0.050 & 0.554 & 0.915 & 23.400 & 0.478 \\
cpo\_opt\_seqlen\_1024\_progressive\_cefr\_parlai\_iteration2 & 6 & 4.812 & 1.341 & 0.067 & 0.543 & 0.922 & 25.200 & 0.488 \\
cpo\_opt\_seqlen\_1024\_progressive\_cefr\_parlai\_iteration3 & 6 & 4.812 & 1.337 & 0.083 & 0.531 & 0.893 & 20.300 & 0.481 \\
cpo\_opt\_seqlen\_1024\_progressive\_cefr\_parlai\_iteration4 & 6 & 4.882 & 1.339 & 0.057 & 0.536 & 0.938 & 22.700 & 0.491 \\
cpo\_opt\_seqlen\_1024\_progressive\_cefr\_parlai\_iteration5 & 6 & 4.886 & 1.333 & 0.064 & 0.528 & 0.929 & 22.600 & 0.490 \\
cpo\_opt\_seqlen\_1024\_progressive\_cefr\_reverse\_parlai\_iteration4 & 6 & 4.846 & 1.331 & 0.036 & 0.557 & 0.916 & 21.300 & 0.450 \\
cpo\_opt\_seqlen\_1024\_progressive\_cefr\_reverse\_parlai\_iteration5 & 6 & 4.982 & 1.373 & 0.069 & 0.541 & 0.853 & 23.000 & 0.471 \\
cpo\_opt\_seqlen\_4096\_final\_checkpoint & 6 & 4.525 & 1.231 & 0.094 & 0.411 & 0.921 & 22.500 & 0.414 \\
mamba-sam-seqlen-2048-original & 6 & 4.881 & 1.325 & 0.059 & 0.558 & 0.969 & 29.200 & 0.522 \\
opt-sam-orpo-mamba-2048-step448 & 6 & 4.831 & 1.369 & 0.084 & 0.553 & 0.912 & 24.200 & 0.514 \\
opt-sam-orpo-seqlen-2048-step559 & 6 & 5.072 & 1.382 & 0.055 & 0.576 & 0.868 & 25.400 & 0.491 \\
opt-sam-seqlen-2048-original & 6 & 4.858 & 1.323 & 0.052 & 0.551 & 0.932 & 24.500 & 0.481 \\
orpo\_opt\_100M\_2048\_preprocess & 6 & 5.145 & 1.411 & 0.082 & 0.546 & 0.865 & 18.400 & 0.522 \\
orpo\_opt\_cosmos & 6 & 5.375 & 1.514 & 0.055 & 0.564 & 0.858 & 17.600 & 0.541 \\
orpo\_opt\_seqlen\_1024\_final\_checkpoint & 6 & 4.951 & 1.390 & 0.065 & 0.567 & 0.927 & 26.100 & 0.526 \\
orpo\_opt\_seqlen\_4096\_final\_checkpoint & 6 & 4.612 & 1.234 & 0.061 & 0.419 & 0.907 & 19.000 & 0.376 \\
babylm-seqlen-opt-1024-warmup-v2 & 6 & 4.864 & 1.339 & 0.038 & 0.543 & 0.901 & 25.300 & 0.446 \\
babylm-seqlen-opt-4096-warmup-v2 & 6 & 4.478 & 1.197 & 0.029 & 0.408 & 0.803 & 22.100 & 0.242 \\
cpo\_opt\_seqlen\_1024\_final\_checkpoint & 8 & 4.917 & 1.346 & 0.058 & 0.523 & 0.893 & 40.100 & 0.476 \\
cpo\_opt\_seqlen\_1024\_progressive\_cefr\_parlai\_iteration1 & 8 & 4.786 & 1.307 & 0.063 & 0.507 & 0.894 & 27.500 & 0.444 \\
cpo\_opt\_seqlen\_1024\_progressive\_cefr\_parlai\_iteration2 & 8 & 5.018 & 1.400 & 0.063 & 0.524 & 0.881 & 30.600 & 0.491 \\
cpo\_opt\_seqlen\_1024\_progressive\_cefr\_parlai\_iteration3 & 8 & 4.790 & 1.324 & 0.066 & 0.489 & 0.842 & 30.500 & 0.413 \\
cpo\_opt\_seqlen\_1024\_progressive\_cefr\_parlai\_iteration4 & 8 & 4.858 & 1.346 & 0.075 & 0.510 & 0.885 & 34.000 & 0.475 \\
cpo\_opt\_seqlen\_1024\_progressive\_cefr\_parlai\_iteration5 & 8 & 4.874 & 1.345 & 0.070 & 0.500 & 0.948 & 29.700 & 0.506 \\
cpo\_opt\_seqlen\_1024\_progressive\_cefr\_reverse\_parlai\_iteration4 & 8 & 4.889 & 1.355 & 0.070 & 0.519 & 0.856 & 29.700 & 0.456 \\
cpo\_opt\_seqlen\_1024\_progressive\_cefr\_reverse\_parlai\_iteration5 & 8 & 4.868 & 1.337 & 0.059 & 0.508 & 0.850 & 30.200 & 0.428 \\
cpo\_opt\_seqlen\_4096\_final\_checkpoint & 8 & 4.495 & 1.193 & 0.073 & 0.363 & 0.941 & 31.100 & 0.378 \\
orpo\_opt\_seqlen\_1024\_final\_checkpoint & 8 & 4.958 & 1.359 & 0.076 & 0.511 & 0.930 & 40.000 & 0.526 \\
orpo\_opt\_seqlen\_4096\_final\_checkpoint & 8 & 4.437 & 1.186 & 0.059 & 0.359 & 0.858 & 32.700 & 0.298 \\
babylm-seqlen-opt-1024-warmup-v2 & 8 & 4.845 & 1.344 & 0.063 & 0.504 & 0.937 & 33.400 & 0.489 \\
babylm-seqlen-opt-4096-warmup-v2 & 8 & 4.458 & 1.170 & 0.036 & 0.337 & 0.897 & 38.300 & 0.290 \\
\bottomrule
\end{tabular}
\caption{Average metrics per BabyLM setting (Length = 50) with min--max normalized aggregate (NormAvg) across metrics.}
\label{tab:dialogue-metrics-len-50}
\end{table*}

\begin{table*}[h!]
\centering
\tiny
\setlength{\tabcolsep}{4pt}
\begin{tabular}{lrrrrrrrr}
\toprule
\textbf{Model} & \textbf{Turns} & \textbf{AoA} & \textbf{CEFR} & \textbf{Overlap} & \textbf{TTR} & \textbf{Rep.} & \textbf{NumCon} & \textbf{NormAvg} \\
\midrule
cpo\_opt\_100M\_2048\_preprocess & 4 & 5.028 & 1.417 & 0.121 & 0.508 & 0.931 & 23.400 & 0.590 \\
cpo\_opt\_base & 4 & 5.075 & 1.404 & 0.088 & 0.496 & 0.847 & 21.000 & 0.491 \\
cpo\_opt\_cosmos & 4 & 5.196 & 1.458 & 0.075 & 0.540 & 0.876 & 25.100 & 0.540 \\
cpo\_opt\_seqlen\_1024\_final\_checkpoint & 4 & 4.921 & 1.353 & 0.062 & 0.537 & 0.955 & 35.900 & 0.524 \\
cpo\_opt\_seqlen\_1024\_progressive\_cefr\_parlai\_iteration1 & 4 & 4.974 & 1.391 & 0.069 & 0.548 & 0.879 & 29.800 & 0.498 \\
cpo\_opt\_seqlen\_1024\_progressive\_cefr\_parlai\_iteration2 & 4 & 4.888 & 1.374 & 0.082 & 0.523 & 0.817 & 31.700 & 0.453 \\
cpo\_opt\_seqlen\_1024\_progressive\_cefr\_parlai\_iteration3 & 4 & 4.900 & 1.365 & 0.083 & 0.522 & 0.903 & 33.600 & 0.511 \\
cpo\_opt\_seqlen\_1024\_progressive\_cefr\_parlai\_iteration4 & 4 & 4.934 & 1.350 & 0.045 & 0.526 & 0.906 & 26.400 & 0.463 \\
cpo\_opt\_seqlen\_1024\_progressive\_cefr\_parlai\_iteration5 & 4 & 4.984 & 1.395 & 0.063 & 0.524 & 0.905 & 25.700 & 0.499 \\
cpo\_opt\_seqlen\_1024\_progressive\_cefr\_reverse\_parlai\_iteration4 & 4 & 4.954 & 1.367 & 0.076 & 0.513 & 0.844 & 21.900 & 0.458 \\
cpo\_opt\_seqlen\_1024\_progressive\_cefr\_reverse\_parlai\_iteration5 & 4 & 4.847 & 1.353 & 0.092 & 0.513 & 0.884 & 27.400 & 0.492 \\
cpo\_opt\_seqlen\_4096\_final\_checkpoint & 4 & 4.532 & 1.224 & 0.058 & 0.391 & 0.891 & 33.300 & 0.349 \\
mamba-sam-seqlen-2048-original & 4 & 4.770 & 1.406 & 0.118 & 0.501 & 0.956 & 40.900 & 0.578 \\
opt-sam-orpo-mamba-2048-step448 & 4 & 4.797 & 1.390 & 0.099 & 0.514 & 0.799 & 42.100 & 0.458 \\
opt-sam-orpo-seqlen-2048-step559 & 4 & 4.728 & 1.304 & 0.077 & 0.512 & 0.893 & 40.100 & 0.463 \\
opt-sam-seqlen-2048-original & 4 & 4.917 & 1.439 & 0.089 & 0.523 & 0.904 & 35.500 & 0.538 \\
orpo\_opt\_100M\_2048\_preprocess & 4 & 5.021 & 1.419 & 0.085 & 0.524 & 0.862 & 20.300 & 0.503 \\
orpo\_opt\_cosmos & 4 & 5.211 & 1.443 & 0.091 & 0.532 & 0.866 & 26.100 & 0.549 \\
orpo\_opt\_seqlen\_1024\_final\_checkpoint & 4 & 4.953 & 1.365 & 0.068 & 0.534 & 0.909 & 35.200 & 0.507 \\
orpo\_opt\_seqlen\_4096\_final\_checkpoint & 4 & 4.711 & 1.262 & 0.062 & 0.419 & 0.870 & 19.600 & 0.371 \\
babylm-seqlen-opt-1024-warmup-v2 & 4 & 4.878 & 1.342 & 0.110 & 0.493 & 0.847 & 38.800 & 0.492 \\
babylm-seqlen-opt-4096-warmup-v2 & 4 & 4.378 & 1.189 & 0.067 & 0.341 & 0.770 & 35.000 & 0.238 \\
cpo\_opt\_100M\_2048\_preprocess & 6 & 5.093 & 1.437 & 0.090 & 0.483 & 0.858 & 36.100 & 0.517 \\
cpo\_opt\_base & 6 & 5.035 & 1.429 & 0.080 & 0.489 & 0.892 & 38.500 & 0.521 \\
cpo\_opt\_cosmos & 6 & 5.075 & 1.395 & 0.116 & 0.465 & 0.904 & 43.300 & 0.566 \\
cpo\_opt\_seqlen\_1024\_final\_checkpoint & 6 & 4.904 & 1.364 & 0.094 & 0.456 & 0.932 & 55.000 & 0.535 \\
cpo\_opt\_seqlen\_1024\_progressive\_cefr\_parlai\_iteration1 & 6 & 4.843 & 1.325 & 0.099 & 0.453 & 0.901 & 51.400 & 0.501 \\
cpo\_opt\_seqlen\_1024\_progressive\_cefr\_parlai\_iteration2 & 6 & 4.805 & 1.323 & 0.092 & 0.460 & 0.892 & 49.000 & 0.482 \\
cpo\_opt\_seqlen\_1024\_progressive\_cefr\_parlai\_iteration3 & 6 & 4.916 & 1.107 & 0.112 & 0.465 & 0.876 & 42.900 & 0.461 \\
cpo\_opt\_seqlen\_1024\_progressive\_cefr\_parlai\_iteration4 & 6 & 4.943 & 1.127 & 0.078 & 0.486 & 0.888 & 40.100 & 0.439 \\
cpo\_opt\_seqlen\_1024\_progressive\_cefr\_parlai\_iteration5 & 6 & 4.845 & 1.366 & 0.095 & 0.469 & 0.912 & 45.600 & 0.514 \\
cpo\_opt\_seqlen\_1024\_progressive\_cefr\_reverse\_parlai\_iteration4 & 6 & 4.953 & 1.363 & 0.068 & 0.469 & 0.915 & 44.500 & 0.495 \\
cpo\_opt\_seqlen\_1024\_progressive\_cefr\_reverse\_parlai\_iteration5 & 6 & 4.974 & 1.394 & 0.075 & 0.483 & 0.858 & 42.800 & 0.478 \\
cpo\_opt\_seqlen\_4096\_final\_checkpoint & 6 & 4.430 & 1.320 & 0.129 & 0.299 & 0.764 & 55.700 & 0.346 \\
mamba-sam-seqlen-2048-original & 6 & 4.838 & 1.429 & 0.105 & 0.474 & 0.840 & 59.100 & 0.503 \\
opt-sam-orpo-mamba-2048-step448 & 6 & 4.959 & 1.411 & 0.066 & 0.478 & 0.962 & 60.700 & 0.549 \\
opt-sam-orpo-seqlen-2048-step559 & 6 & 4.912 & 1.352 & 0.080 & 0.508 & 0.944 & 44.600 & 0.535 \\
opt-sam-seqlen-2048-original & 6 & 4.861 & 1.384 & 0.071 & 0.465 & 0.838 & 52.600 & 0.446 \\
orpo\_opt\_100M\_2048\_preprocess & 6 & 4.984 & 1.399 & 0.093 & 0.479 & 0.853 & 41.800 & 0.498 \\
orpo\_opt\_cosmos & 6 & 5.155 & 1.512 & 0.111 & 0.509 & 0.917 & 38.800 & 0.616 \\
orpo\_opt\_seqlen\_1024\_final\_checkpoint & 6 & 4.965 & 1.382 & 0.069 & 0.467 & 0.888 & 57.200 & 0.492 \\
orpo\_opt\_seqlen\_4096\_final\_checkpoint & 6 & 4.626 & 1.274 & 0.060 & 0.362 & 0.937 & 34.500 & 0.396 \\
babylm-seqlen-opt-1024-warmup-v2 & 6 & 4.895 & 1.365 & 0.080 & 0.467 & 0.955 & 55.600 & 0.536 \\
babylm-seqlen-opt-4096-warmup-v2 & 6 & 4.709 & 1.256 & 0.049 & 0.317 & 0.842 & 48.900 & 0.320 \\
cpo\_opt\_seqlen\_1024\_final\_checkpoint & 8 & 4.956 & 1.406 & 0.113 & 0.421 & 0.953 & 69.300 & 0.584 \\
cpo\_opt\_seqlen\_1024\_progressive\_cefr\_parlai\_iteration1 & 8 & 4.938 & 1.389 & 0.081 & 0.436 & 0.863 & 69.200 & 0.485 \\
cpo\_opt\_seqlen\_1024\_progressive\_cefr\_parlai\_iteration2 & 8 & 4.954 & 1.402 & 0.083 & 0.435 & 0.898 & 63.500 & 0.512 \\
cpo\_opt\_seqlen\_1024\_progressive\_cefr\_parlai\_iteration3 & 8 & 5.062 & 1.417 & 0.090 & 0.440 & 0.917 & 63.100 & 0.551 \\
cpo\_opt\_seqlen\_1024\_progressive\_cefr\_parlai\_iteration4 & 8 & 4.939 & 1.116 & 0.102 & 0.437 & 0.893 & 60.400 & 0.466 \\
cpo\_opt\_seqlen\_1024\_progressive\_cefr\_parlai\_iteration5 & 8 & 5.110 & 1.419 & 0.084 & 0.431 & 0.930 & 55.600 & 0.550 \\
cpo\_opt\_seqlen\_1024\_progressive\_cefr\_reverse\_parlai\_iteration4 & 8 & 5.052 & 1.159 & 0.085 & 0.448 & 0.940 & 56.200 & 0.500 \\
cpo\_opt\_seqlen\_1024\_progressive\_cefr\_reverse\_parlai\_iteration5 & 8 & 4.903 & 1.104 & 0.108 & 0.431 & 0.879 & 58.300 & 0.453 \\
cpo\_opt\_seqlen\_4096\_final\_checkpoint & 8 & 4.493 & 1.187 & 0.073 & 0.273 & 0.832 & 89.500 & 0.313 \\
orpo\_opt\_seqlen\_1024\_final\_checkpoint & 8 & 5.052 & 1.424 & 0.072 & 0.443 & 0.957 & 74.700 & 0.564 \\
orpo\_opt\_seqlen\_4096\_final\_checkpoint & 8 & 4.482 & 1.315 & 0.120 & 0.294 & 0.903 & 51.900 & 0.426 \\
babylm-seqlen-opt-1024-warmup-v2 & 8 & 5.002 & 1.432 & 0.084 & 0.447 & 0.829 & 59.200 & 0.482 \\
babylm-seqlen-opt-4096-warmup-v2 & 8 & 4.485 & 1.215 & 0.066 & 0.279 & 0.914 & 70.000 & 0.352 \\
\bottomrule
\end{tabular}
\caption{Average metrics per BabyLM setting (Length = 100) with min--max normalized aggregate (NormAvg) across metrics.}
\label{tab:dialogue-metrics-len-100}
\end{table*}

\begin{table*}[t]
\centering
\tiny
\setlength{\tabcolsep}{4pt}
\begin{tabular}{lrrrrrrrr}
\toprule
\textbf{Model} & \textbf{Turns} & \textbf{AoA} & \textbf{CEFR} & \textbf{Overlap} & \textbf{TTR} & \textbf{Rep.} & \textbf{NumCon} & \textbf{NormAvg} \\
\midrule
cpo\_opt\_100M\_2048\_preprocess & 4 & 5.057 & 1.441 & 0.115 & 0.469 & 0.828 & 33.400 & 0.518 \\
cpo\_opt\_base & 4 & 5.153 & 1.433 & 0.082 & 0.458 & 0.883 & 34.500 & 0.519 \\
cpo\_opt\_cosmos & 4 & 5.174 & 1.460 & 0.076 & 0.488 & 0.894 & 33.800 & 0.539 \\
cpo\_opt\_seqlen\_1024\_final\_checkpoint & 4 & 4.912 & 1.360 & 0.104 & 0.445 & 0.959 & 52.600 & 0.560 \\
cpo\_opt\_seqlen\_1024\_progressive\_cefr\_parlai\_iteration1 & 4 & 4.893 & 1.360 & 0.092 & 0.469 & 0.896 & 43.200 & 0.502 \\
cpo\_opt\_seqlen\_1024\_progressive\_cefr\_parlai\_iteration2 & 4 & 4.923 & 1.418 & 0.105 & 0.467 & 0.936 & 45.100 & 0.562 \\
cpo\_opt\_seqlen\_1024\_progressive\_cefr\_parlai\_iteration3 & 4 & 4.949 & 1.353 & 0.109 & 0.439 & 0.901 & 40.200 & 0.519 \\
cpo\_opt\_seqlen\_1024\_progressive\_cefr\_parlai\_iteration4 & 4 & 5.055 & 1.350 & 0.083 & 0.448 & 0.889 & 33.300 & 0.490 \\
cpo\_opt\_seqlen\_1024\_progressive\_cefr\_parlai\_iteration5 & 4 & 4.989 & 1.379 & 0.114 & 0.443 & 0.870 & 34.600 & 0.514 \\
cpo\_opt\_seqlen\_1024\_progressive\_cefr\_reverse\_parlai\_iteration4 & 4 & 4.983 & 1.375 & 0.081 & 0.499 & 0.894 & 35.000 & 0.507 \\
cpo\_opt\_seqlen\_1024\_progressive\_cefr\_reverse\_parlai\_iteration5 & 4 & 4.901 & 1.380 & 0.070 & 0.470 & 0.862 & 40.900 & 0.457 \\
cpo\_opt\_seqlen\_4096\_final\_checkpoint & 4 & 4.385 & 1.256 & 0.128 & 0.302 & 0.749 & 49.200 & 0.311 \\
mamba-sam-seqlen-2048-original & 4 & 4.918 & 1.435 & 0.075 & 0.470 & 0.906 & 55.800 & 0.516 \\
opt-sam-orpo-mamba-2048-step448 & 4 & 4.880 & 1.384 & 0.073 & 0.470 & 0.870 & 64.800 & 0.481 \\
opt-sam-orpo-seqlen-2048-step559 & 4 & 4.910 & 1.371 & 0.069 & 0.473 & 0.921 & 47.300 & 0.500 \\
opt-sam-seqlen-2048-original & 4 & 4.998 & 1.416 & 0.117 & 0.466 & 0.913 & 45.500 & 0.570 \\
orpo\_opt\_100M\_2048\_preprocess & 4 & 5.169 & 1.432 & 0.109 & 0.473 & 0.766 & 33.300 & 0.483 \\
orpo\_opt\_cosmos & 4 & 5.284 & 1.489 & 0.102 & 0.507 & 0.847 & 27.000 & 0.561 \\
orpo\_opt\_seqlen\_1024\_final\_checkpoint & 4 & 5.077 & 1.395 & 0.076 & 0.474 & 0.825 & 69.600 & 0.487 \\
orpo\_opt\_seqlen\_4096\_final\_checkpoint & 4 & 4.732 & 1.301 & 0.078 & 0.391 & 0.855 & 37.700 & 0.395 \\
babylm-seqlen-opt-1024-warmup-v2 & 4 & 4.778 & 1.351 & 0.098 & 0.441 & 0.903 & 55.800 & 0.497 \\
babylm-seqlen-opt-4096-warmup-v2 & 4 & 4.473 & 1.222 & 0.032 & 0.355 & 0.900 & 50.300 & 0.316 \\
cpo\_opt\_seqlen\_1024\_final\_checkpoint & 6 & 5.236 & 1.493 & 0.075 & 0.460 & 0.954 & 75.800 & 0.610 \\
cpo\_opt\_seqlen\_1024\_progressive\_cefr\_parlai\_iteration1 & 6 & 4.859 & 1.371 & 0.090 & 0.414 & 0.912 & 76.200 & 0.513 \\
cpo\_opt\_seqlen\_1024\_progressive\_cefr\_parlai\_iteration2 & 6 & 5.048 & 1.427 & 0.088 & 0.438 & 0.918 & 60.700 & 0.547 \\
cpo\_opt\_seqlen\_1024\_progressive\_cefr\_parlai\_iteration3 & 6 & 5.091 & 0.960 & 0.086 & 0.420 & 0.897 & 62.100 & 0.428 \\
cpo\_opt\_seqlen\_1024\_progressive\_cefr\_parlai\_iteration4 & 6 & 5.128 & 1.419 & 0.088 & 0.420 & 0.870 & 71.100 & 0.525 \\
cpo\_opt\_seqlen\_1024\_progressive\_cefr\_parlai\_iteration5 & 6 & 5.011 & 1.397 & 0.092 & 0.429 & 0.902 & 67.300 & 0.532 \\
cpo\_opt\_seqlen\_1024\_progressive\_cefr\_reverse\_parlai\_iteration4 & 6 & 5.152 & 1.473 & 0.093 & 0.450 & 0.888 & 66.200 & 0.564 \\
cpo\_opt\_seqlen\_1024\_progressive\_cefr\_reverse\_parlai\_iteration5 & 6 & 4.943 & 1.375 & 0.165 & 0.390 & 0.908 & 64.800 & 0.597 \\
cpo\_opt\_seqlen\_4096\_final\_checkpoint & 6 & 4.410 & 1.191 & 0.070 & 0.252 & 0.885 & 78.400 & 0.320 \\
orpo\_opt\_seqlen\_1024\_final\_checkpoint & 6 & 5.056 & 1.165 & 0.086 & 0.437 & 0.827 & 75.800 & 0.439 \\
orpo\_opt\_seqlen\_4096\_final\_checkpoint & 6 & 4.621 & 1.283 & 0.077 & 0.306 & 0.876 & 58.000 & 0.374 \\
babylm-seqlen-opt-1024-warmup-v2 & 6 & 5.060 & 1.429 & 0.076 & 0.439 & 0.858 & 66.000 & 0.499 \\
babylm-seqlen-opt-4096-warmup-v2 & 6 & 4.537 & 1.213 & 0.063 & 0.266 & 0.890 & 89.300 & 0.347 \\
cpo\_opt\_seqlen\_4096\_final\_checkpoint & 8 & 4.408 & 1.170 & 0.068 & 0.201 & 0.900 & 115.300 & 0.330 \\
orpo\_opt\_seqlen\_4096\_final\_checkpoint & 8 & 4.481 & 1.202 & 0.070 & 0.248 & 0.890 & 85.200 & 0.338 \\
babylm-seqlen-opt-4096-warmup-v2 & 8 & 4.505 & 1.218 & 0.036 & 0.251 & 0.856 & 100.000 & 0.293 \\
\bottomrule
\end{tabular}
\caption{Average metrics per BabyLM setting (Length = 150) with min--max normalized aggregate (NormAvg) across metrics.}
\label{tab:dialogue-metrics-len-150}
\end{table*}

\begin{table*}[t]
\centering
\tiny
\setlength{\tabcolsep}{4pt}
\begin{tabular}{lrrrrrrrr}
\toprule
\textbf{Model} & \textbf{Turns} & \textbf{AoA} & \textbf{CEFR} & \textbf{Overlap} & \textbf{TTR} & \textbf{Rep.} & \textbf{NumCon} & \textbf{NormAvg} \\
\midrule
cpo\_opt\_100M\_2048\_preprocess & 4 & 5.171 & 0.932 & 0.108 & 0.432 & 0.904 & 40.300 & 0.453 \\
cpo\_opt\_base & 4 & 5.378 & 1.506 & 0.099 & 0.433 & 0.882 & 41.500 & 0.580 \\
cpo\_opt\_cosmos & 4 & 5.191 & 1.270 & 0.130 & 0.435 & 0.933 & 50.800 & 0.585 \\
cpo\_opt\_seqlen\_1024\_final\_checkpoint & 4 & 4.960 & 1.394 & 0.105 & 0.427 & 0.891 & 67.500 & 0.533 \\
cpo\_opt\_seqlen\_1024\_progressive\_cefr\_parlai\_iteration1 & 4 & 4.827 & 1.345 & 0.113 & 0.430 & 0.946 & 59.900 & 0.547 \\
cpo\_opt\_seqlen\_1024\_progressive\_cefr\_parlai\_iteration2 & 4 & 4.952 & 1.391 & 0.061 & 0.419 & 0.874 & 58.100 & 0.457 \\
cpo\_opt\_seqlen\_1024\_progressive\_cefr\_parlai\_iteration3 & 4 & 5.184 & 1.474 & 0.115 & 0.434 & 0.873 & 60.400 & 0.576 \\
cpo\_opt\_seqlen\_1024\_progressive\_cefr\_parlai\_iteration4 & 4 & 4.899 & 1.337 & 0.071 & 0.469 & 0.872 & 44.300 & 0.458 \\
cpo\_opt\_seqlen\_1024\_progressive\_cefr\_parlai\_iteration5 & 4 & 5.025 & 1.387 & 0.093 & 0.441 & 0.904 & 50.500 & 0.527 \\
cpo\_opt\_seqlen\_1024\_progressive\_cefr\_reverse\_parlai\_iteration4 & 4 & 5.006 & 1.444 & 0.068 & 0.454 & 0.887 & 43.300 & 0.494 \\
cpo\_opt\_seqlen\_1024\_progressive\_cefr\_reverse\_parlai\_iteration5 & 4 & 5.163 & 1.257 & 0.083 & 0.453 & 0.875 & 50.100 & 0.488 \\
cpo\_opt\_seqlen\_4096\_final\_checkpoint & 4 & 4.367 & 1.188 & 0.082 & 0.245 & 0.770 & 102.700 & 0.269 \\
mamba-sam-seqlen-2048-original & 4 & 4.967 & 1.547 & 0.108 & 0.430 & 0.819 & 73.800 & 0.529 \\
opt-sam-orpo-mamba-2048-step448 & 4 & 5.049 & 1.435 & 0.095 & 0.438 & 0.853 & 76.100 & 0.525 \\
opt-sam-orpo-seqlen-2048-step559 & 4 & 4.825 & 1.152 & 0.090 & 0.456 & 0.829 & 62.500 & 0.410 \\
opt-sam-seqlen-2048-original & 4 & 4.947 & 1.384 & 0.073 & 0.456 & 0.841 & 59.800 & 0.462 \\
orpo\_opt\_100M\_2048\_preprocess & 4 & 5.080 & 1.198 & 0.100 & 0.463 & 0.928 & 39.000 & 0.516 \\
orpo\_opt\_cosmos & 4 & 5.417 & 1.546 & 0.074 & 0.478 & 0.854 & 43.100 & 0.562 \\
orpo\_opt\_seqlen\_1024\_final\_checkpoint & 4 & 5.047 & 0.888 & 0.094 & 0.438 & 0.948 & 71.000 & 0.463 \\
orpo\_opt\_seqlen\_4096\_final\_checkpoint & 4 & 4.533 & 1.340 & 0.122 & 0.306 & 0.808 & 65.100 & 0.391 \\
babylm-seqlen-opt-1024-warmup-v2 & 4 & 5.010 & 1.410 & 0.072 & 0.436 & 0.952 & 75.800 & 0.550 \\
babylm-seqlen-opt-4096-warmup-v2 & 4 & 4.418 & 1.164 & 0.042 & 0.268 & 0.750 & 70.400 & 0.193 \\
cpo\_opt\_seqlen\_1024\_progressive\_cefr\_parlai\_iteration1 & 6 & 5.023 & 1.419 & 0.089 & 0.412 & 0.898 & 91.200 & 0.542 \\
cpo\_opt\_seqlen\_1024\_progressive\_cefr\_parlai\_iteration2 & 6 & 4.785 & 1.297 & 0.122 & 0.363 & 0.776 & 117.700 & 0.449 \\
cpo\_opt\_seqlen\_1024\_progressive\_cefr\_parlai\_iteration3 & 6 & 4.934 & 1.398 & 0.120 & 0.369 & 0.832 & 88.800 & 0.505 \\
cpo\_opt\_seqlen\_1024\_progressive\_cefr\_parlai\_iteration4 & 6 & 4.947 & 1.363 & 0.142 & 0.378 & 0.888 & 79.700 & 0.560 \\
cpo\_opt\_seqlen\_1024\_progressive\_cefr\_parlai\_iteration5 & 6 & 4.964 & 1.372 & 0.090 & 0.411 & 0.889 & 71.100 & 0.506 \\
cpo\_opt\_seqlen\_1024\_progressive\_cefr\_reverse\_parlai\_iteration5 & 6 & 5.100 & 1.431 & 0.088 & 0.382 & 0.915 & 77.400 & 0.544 \\
cpo\_opt\_seqlen\_4096\_final\_checkpoint & 6 & 4.510 & 1.267 & 0.084 & 0.244 & 0.809 & 98.300 & 0.328 \\
orpo\_opt\_seqlen\_4096\_final\_checkpoint & 6 & 4.456 & 1.293 & 0.120 & 0.245 & 0.909 & 118.600 & 0.451 \\
babylm-seqlen-opt-1024-warmup-v2 & 6 & 4.897 & 1.397 & 0.110 & 0.376 & 0.946 & 100.700 & 0.573 \\
babylm-seqlen-opt-4096-warmup-v2 & 6 & 4.402 & 1.181 & 0.047 & 0.227 & 0.870 & 113.300 & 0.295 \\
cpo\_opt\_seqlen\_4096\_final\_checkpoint & 8 & 4.366 & 1.223 & 0.109 & 0.196 & 0.934 & 198.200 & 0.464 \\
orpo\_opt\_seqlen\_4096\_final\_checkpoint & 8 & 4.663 & 1.284 & 0.107 & 0.247 & 0.830 & 113.100 & 0.403 \\
babylm-seqlen-opt-4096-warmup-v2 & 8 & 4.446 & 1.166 & 0.114 & 0.183 & 0.841 & 154.400 & 0.372 \\
\bottomrule
\end{tabular}
\caption{Average metrics per BabyLM setting (Length = 200) with min--max normalized aggregate (NormAvg) across metrics.}
\label{tab:dialogue-metrics-len-200}
\end{table*}

\begin{table*}[t]
\centering
\tiny
\setlength{\tabcolsep}{4pt}
\begin{tabular}{lrrrrrrrr}
\toprule
\textbf{Model} & \textbf{Turns} & \textbf{AoA} & \textbf{CEFR} & \textbf{Overlap} & \textbf{TTR} & \textbf{Rep.} & \textbf{NumCon} & \textbf{NormAvg} \\
\midrule
cpo\_opt\_100M\_2048\_preprocess & 4 & 5.054 & 1.393 & 0.125 & 0.396 & 0.950 & 57.700 & 0.590 \\
cpo\_opt\_base & 4 & 5.655 & 1.647 & 0.108 & 0.411 & 0.835 & 47.300 & 0.622 \\
cpo\_opt\_cosmos & 4 & 5.124 & 1.435 & 0.104 & 0.419 & 0.855 & 61.500 & 0.531 \\
cpo\_opt\_seqlen\_1024\_final\_checkpoint & 4 & 4.932 & 1.391 & 0.103 & 0.411 & 0.819 & 80.200 & 0.482 \\
cpo\_opt\_seqlen\_1024\_progressive\_cefr\_parlai\_iteration1 & 4 & 5.149 & 1.467 & 0.104 & 0.399 & 0.910 & 60.400 & 0.569 \\
cpo\_opt\_seqlen\_1024\_progressive\_cefr\_parlai\_iteration2 & 4 & 4.927 & 1.378 & 0.099 & 0.396 & 0.840 & 55.800 & 0.465 \\
cpo\_opt\_seqlen\_1024\_progressive\_cefr\_parlai\_iteration3 & 4 & 4.898 & 1.375 & 0.140 & 0.409 & 0.805 & 60.800 & 0.497 \\
cpo\_opt\_seqlen\_1024\_progressive\_cefr\_parlai\_iteration4 & 4 & 5.029 & 1.427 & 0.092 & 0.387 & 0.887 & 66.900 & 0.516 \\
cpo\_opt\_seqlen\_1024\_progressive\_cefr\_parlai\_iteration5 & 4 & 5.183 & 1.434 & 0.095 & 0.419 & 0.838 & 59.000 & 0.514 \\
cpo\_opt\_seqlen\_1024\_progressive\_cefr\_reverse\_parlai\_iteration4 & 4 & 5.007 & 1.391 & 0.101 & 0.409 & 0.839 & 48.600 & 0.480 \\
cpo\_opt\_seqlen\_1024\_progressive\_cefr\_reverse\_parlai\_iteration5 & 4 & 5.016 & 0.910 & 0.155 & 0.370 & 0.911 & 68.100 & 0.489 \\
cpo\_opt\_seqlen\_4096\_final\_checkpoint & 4 & 4.632 & 1.245 & 0.094 & 0.286 & 0.891 & 78.300 & 0.405 \\
mamba-sam-seqlen-2048-original & 4 & 4.836 & 1.413 & 0.100 & 0.411 & 0.869 & 100.300 & 0.519 \\
opt-sam-orpo-mamba-2048-step448 & 4 & 4.777 & 1.140 & 0.126 & 0.388 & 0.778 & 108.300 & 0.421 \\
opt-sam-orpo-seqlen-2048-step559 & 4 & 5.091 & 1.454 & 0.092 & 0.407 & 0.861 & 67.500 & 0.520 \\
opt-sam-seqlen-2048-original & 4 & 4.997 & 1.439 & 0.116 & 0.401 & 0.907 & 71.200 & 0.565 \\
orpo\_opt\_100M\_2048\_preprocess & 4 & 5.160 & 1.247 & 0.101 & 0.424 & 0.897 & 47.700 & 0.510 \\
orpo\_opt\_cosmos & 4 & 5.386 & 1.497 & 0.120 & 0.436 & 0.844 & 47.600 & 0.585 \\
orpo\_opt\_seqlen\_1024\_final\_checkpoint & 4 & 5.073 & 1.197 & 0.096 & 0.412 & 0.888 & 90.700 & 0.502 \\
orpo\_opt\_seqlen\_4096\_final\_checkpoint & 4 & 4.741 & 1.391 & 0.117 & 0.318 & 0.810 & 49.500 & 0.416 \\
babylm-seqlen-opt-1024-warmup-v2 & 4 & 4.977 & 1.426 & 0.077 & 0.429 & 0.905 & 71.000 & 0.520 \\
babylm-seqlen-opt-4096-warmup-v2 & 4 & 4.549 & 1.223 & 0.070 & 0.266 & 0.922 & 79.400 & 0.374 \\
cpo\_opt\_seqlen\_4096\_final\_checkpoint & 6 & 4.375 & 1.179 & 0.104 & 0.212 & 0.857 & 132.200 & 0.360 \\
orpo\_opt\_seqlen\_4096\_final\_checkpoint & 6 & 4.498 & 1.226 & 0.151 & 0.214 & 0.932 & 130.800 & 0.492 \\
babylm-seqlen-opt-4096-warmup-v2 & 6 & 4.298 & 1.118 & 0.070 & 0.168 & 0.780 & 182.400 & 0.264 \\
cpo\_opt\_seqlen\_4096\_final\_checkpoint & 8 & 4.366 & 1.207 & 0.129 & 0.159 & 0.838 & 245.300 & 0.442 \\
orpo\_opt\_seqlen\_4096\_final\_checkpoint & 8 & 4.434 & 1.283 & 0.141 & 0.181 & 0.938 & 254.000 & 0.561 \\
babylm-seqlen-opt-4096-warmup-v2 & 8 & 4.531 & 1.193 & 0.063 & 0.171 & 0.715 & 162.200 & 0.245 \\
\bottomrule
\end{tabular}
\caption{Average metrics per BabyLM setting (Length = 250) with min--max normalized aggregate (NormAvg) across metrics.}
\label{tab:dialogue-metrics-len-250}
\end{table*}

%\end{landscape}

\clearpage
\subsection{Effect of Meta-Prompts}

We additionally report the effect of providing meta-prompts (using the template in \textit{Section} \ref{meta-prompt}) to the Teacher Model on our automatic metrics. We report results for our best performing student model \verb|cpo-opt-1024|.

\begin{table*}[ht]
\centering
\small
\begin{tabular*}{\linewidth}{lccccccccc}
\toprule
\textbf{Age} & \textbf{TurnNo.} & \textbf{AvgSentLen} & \textbf{AoA} & \textbf{AddCon} & \textbf{CEFR} & \textbf{Overlap} & \textbf{CausalCon} & \textbf{TTR} & \textbf{ConceptDensity} \\ \midrule
\multirow{4}{*}{6-11m} & 1 & 21.304 & 5.080 & 3.600 & 1.390 & 0.201 & 0.700 & 0.483 & 5.399 \\
 & 3 & 25.919 & 5.008 & 7.350 & 1.381 & 0.250 & 1.700 & 0.232 & 5.613 \\
 & 5 & 29.969 & 4.940 & 10.500 & 1.381 & 0.286 & 3.050 & 0.163 & 5.666  \\
 & all & 34.947 & 4.860 & 15.500 & 1.354 & 0.324 & 4.050 & 0.107 & 5.886  \\
\midrule
\multirow{4}{*}{18-23m} & 1 & 15.193 & 4.734 & 2.750 & 1.296 & 0.307 & 0.900 & 0.454 & 4.213  \\
 & 3 & 17.580 & 4.689 & 6.600 & 1.277 & 0.330 & 2.300 & 0.234 & 4.450  \\
 & 5 & 17.818 & 4.689 & 10.900 & 1.272 & 0.352 & 3.750 & 0.164 & 4.422  \\
 & all & 20.676 & 4.609 & 16.050 & 1.241 & 0.411 & 5.150 & 0.099 & 4.332 \\
\midrule
\multirow{4}{*}{2-3y} & 1 & 17.008 & 4.534 & 3.550 & 1.209 & 0.112 & 0.400 & 0.554 & 4.039  \\
& 3 & 18.100 & 4.460 & 7.600 & 1.197 & 0.178 & 1.100 & 0.330 & 4.155 \\
& 5 & 18.934 & 4.428 & 11.350 & 1.192 & 0.202 & 1.800 & 0.246 & 4.153  \\
 & all & 22.438 & 4.390 & 18.400 & 1.159 & 0.240 & 2.550 & 0.150 & 4.302 \\
\midrule
\multirow{4}{*}{3-4y} & 1 & 16.645 & 5.446 & 2.900 & 1.500 & 0.182 & 0.000 & 0.539 & 4.984 \\
 & 3 & 18.344 & 5.348 & 6.450 & 1.485 & 0.251 & 0.100 & 0.297 & 5.000 \\ 
 & 5 & 19.668 & 5.357 & 9.200 & 1.493 & 0.296 & 0.200 & 0.211 & 4.990  \\
 & all & 23.885 & 5.270 & 16.050 & 1.449 & 0.360 & 0.700 & 0.114 & 5.127 \\
\midrule
\multirow{4}{*}{4-5y} & 1 & 27.381 & 4.943 & 2.850 & 1.350 & 0.260 & 0.250 & 0.443 & 5.096  \\
& 3 & 25.607 & 4.880 & 9.000 & 1.315 & 0.203 & 2.000 & 0.295 & 5.094 \\
 & 5 & 27.058 & 4.842 & 14.950 & 1.306 & 0.183 & 3.250 & 0.223 & 5.072  \\
& all & 31.231 & 4.778 & 23.000 & 1.280 & 0.191 & 4.600 & 0.146 & 5.300 \\
\bottomrule
\end{tabular*}
\caption{Average metrics by age (where ``m'' is short for ``months'' and ``y'' is short for ``years'') and number of Student-Teacher Turns (\textbf{TurnNo.}; ordered as 1, 3, 5, all). Normalized average uses min–max normalization across model outputs per metric. Note that this table is continued below and on the next page to accommodate all of the linguistic complexity metrics we measured.}
\label{tab:age-segment-metrics}
\end{table*}

\setcounter{table}{14}
\renewcommand{\thetable}{\arabic{table} (contd.)}

\begin{table*}[h!]
\centering
\small
\begin{tabular*}{\linewidth}{lcccccccc}
\toprule
\textbf{Age} & \textbf{TurnNo.} & \textbf{VerbOverlap} & \textbf{AvgClauses} & \textbf{MATTR} & \textbf{AvgFam} & \textbf{Rep.} & \textbf{NumCon} & \textbf{AdversativeCon} \\ \midrule
\multirow{4}{*}{6-11m} & 1 & 0.152 & 2.316 & 0.629 & 13.563 & 0.844 & 4.350 & 0.500 \\
 & 3 & 0.193 & 2.301 & 0.583 & 13.638 & 0.837 & 9.300 & 1.400  \\
 & 5 & 0.206 & 2.373 & 0.563 & 13.704 & 0.815 & 14.000 & 2.050  \\
 & all & 0.230 & 2.267 & 0.517 & 13.756 & 0.849 & 20.150 & 2.800\\
\midrule
\multirow{4}{*}{18-23m} & 1 & 0.171 & 1.833 & 0.600 & 13.597 & 0.659 & 3.750 & 0.700  \\
 & 3 & 0.188 & 1.958 & 0.548 & 13.639 & 0.698 & 9.300 & 1.900  \\
 & 5 &  0.188 & 1.926 & 0.534 & 13.604 & 0.732 & 15.500 & 3.250  \\
 & all  & 0.209 & 1.902 & 0.485 & 13.617 & 0.766 & 22.200 & 5.150  \\
\midrule
\multirow{4}{*}{2-3y} & 1 & 0.101 & 2.220 & 0.691 & 14.155 & 0.674 & 4.800 & 1.200 \\
& 3 &  0.146 & 2.244 & 0.642 & 14.260 & 0.734 & 10.400 & 2.650  \\
& 5 & 0.175 & 2.302 & 0.624 & 14.329 & 0.776 & 16.000 & 3.500  \\
 & all  & 0.234 & 2.444 & 0.561 & 14.232 & 0.833 & 26.500 & 4.850 \\
\midrule
\multirow{4}{*}{3-4y} & 1  & 0.146 & 1.697 & 0.666 & 13.255 & 0.818 & 3.200 & 0.450  \\
 & 3 & 0.183 & 1.826 & 0.593 & 13.310 & 0.821 & 7.350 & 1.300  \\ 
 & 5 & 0.207 & 1.959 & 0.567 & 13.335 & 0.838 & 10.800 & 1.850  \\
 & all & 0.238 & 1.799 & 0.515 & 13.271 & 0.878 & 20.900 & 3.500  \\
\midrule
\multirow{4}{*}{4-5y} & 1 & 0.249 & 2.811 & 0.564 & 13.331 & 0.650 & 3.450 & 0.650  \\
& 3 & 0.159 & 2.496 & 0.580 & 13.736 & 0.878 & 11.550 & 1.700  \\
 & 5 & 0.146 & 2.676 & 0.589 & 13.844 & 0.869 & 18.800 & 2.600 \\
& all  & 0.127 & 2.684 & 0.546 & 13.776 & 0.858 & 29.100 & 3.850  \\
\bottomrule
\end{tabular*}
\caption{Average metrics by \textbf{Age} (where ``m'' is short for ``months'' and ``y'' is short for ``years'') and number of Student-Teacher Turns (\textbf{TurnNo.}; ordered as 1, 3, 5, all). Normalized average uses min–max normalization across model outputs per metric.}
\label{tab:age-segment-metrics-cont}
\end{table*}

\setcounter{table}{14}

\begin{table*}[ht]
\centering
\small
\begin{tabular*}{\linewidth}{lccccc}
\toprule
\textbf{Age} & \textbf{TurnNo.} & \textbf{Polysemy} & \textbf{VerbRep} & \textbf{Narrativity} & \textbf{Norm. Avg} \\ \midrule
\multirow{4}{*}{6-11m} & 1 & 8.826 & 0.492 & 0.000 & 0.438 \\
 & 3 & 8.917 & 0.632 & 0.000 & 0.476 \\
 & 5 & 8.741 & 0.688 & 0.000 & 0.556 \\
 & all & 8.585 & 0.702 & 0.000 & 0.611 \\
\midrule
\multirow{4}{*}{18-23m} & 1 & 10.066 & 0.636 & -0.000 & 0.299 \\
 & 3 & 10.278 & 0.706 & 0.000 & 0.382 \\
 & 5 & 10.090 & 0.703 & 0.000 & 0.443 \\
 & all  & 10.067 & 0.731 & -0.000 & 0.484 \\
\midrule
\multirow{4}{*}{2-3y} & 1 & 10.200 & 0.854 & -0.000 & 0.339 \\
& 3 & 10.706 & 0.809 & -0.000 & 0.388 \\
& 5 & 10.976 & 0.803 & -0.000 & 0.457 \\
 & all  &  10.844 & 0.817 & -0.000 & 0.561 \\
\midrule
\multirow{4}{*}{3-4y} & 1  & 7.821 & 0.562 & -0.000 & 0.354 \\
 & 3 &  7.651 & 0.617 & 0.000 & 0.393 \\ 
 & 5 &  7.514 & 0.637 & 0.000 & 0.437 \\
 & all & 7.403 & 0.669 & 0.000 & 0.504 \\
\midrule
\multirow{4}{*}{4-5y} & 1 &  8.411 & 0.629 & -0.000 & 0.398 \\
& 3 & 8.819 & 0.726 & 0.000 & 0.489 \\
 & 5 & 9.207 & 0.681 & -0.000 & 0.526 \\
& all  & 9.084 & 0.662 & -0.000 & 0.573 \\
\bottomrule
\end{tabular*}
\caption{Average metrics by age (where ``m'' is short for ``months'' and ``y'' is short for ``years'') and number of Student-Teacher Turns (\textbf{TurnNo.}; ordered as 1, 3, 5, all). Normalized average uses min–max normalization across model outputs per metric.}
\label{tab:age-segment-metrics-cont-cont}
\end{table*}

\renewcommand{\thetable}{\arabic{table}}

\clearpage 
\section{Analysis of Reward Model Training Dynamics for CPO/ORPO and CEFR Models} \label{training-metrics}
\subsection{Comparison of Training Reward Dynamics Across Reward Types for CPO/ORPO (Experiment 1) and CEFR Models (Experiment 2)}

\begin{figure*}[h!]
    \centering
    \includegraphics[width=\textwidth]{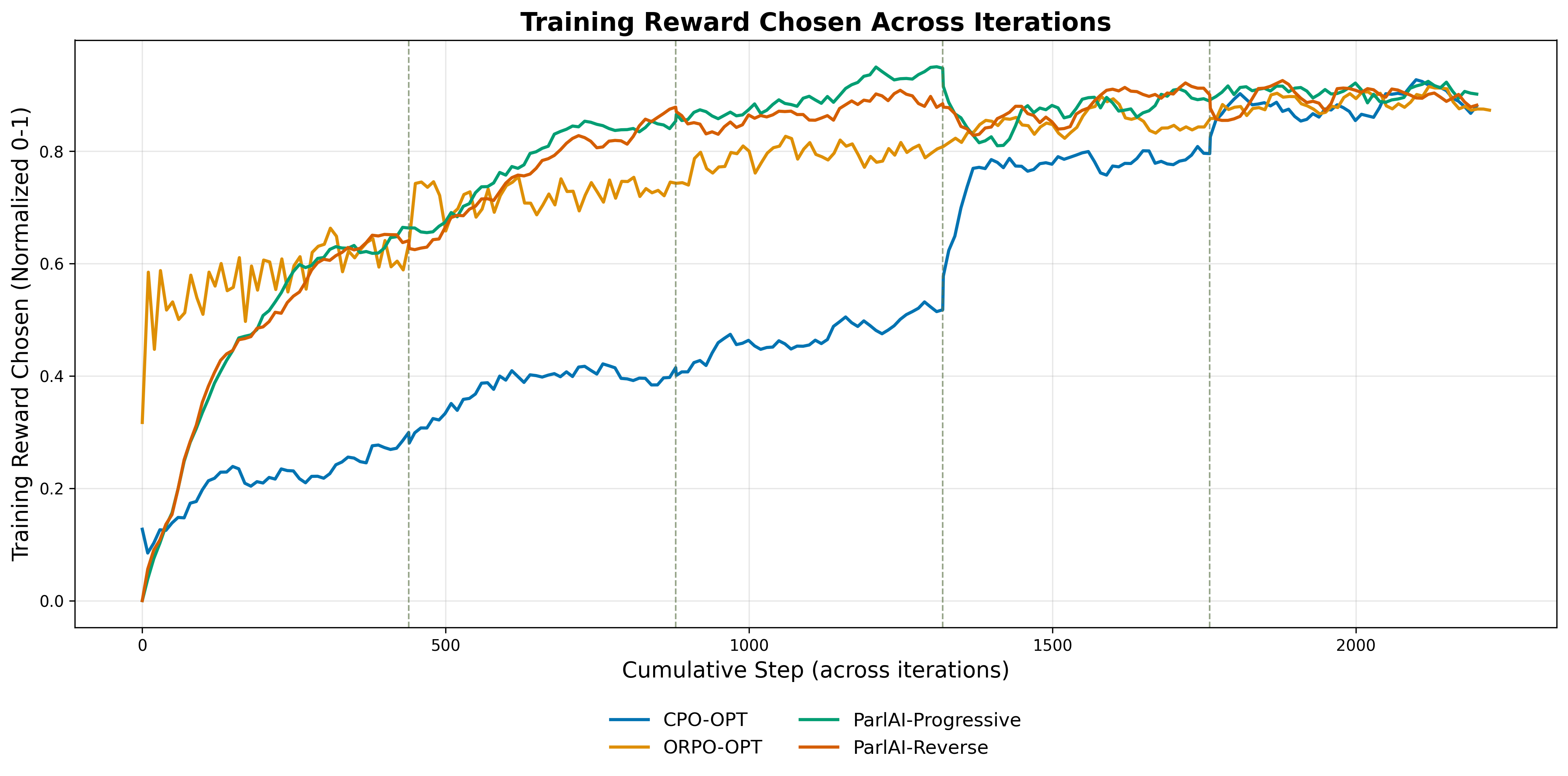}
    \caption{\name Training Reward Chosen for CPO/ORPO Models (Experiment 1) and CEFR Models (Experiment 2)}
    \label{fig:reward_chosen_1}
\end{figure*}

\begin{figure*}[h!]
    \centering
    \includegraphics[width=\textwidth]{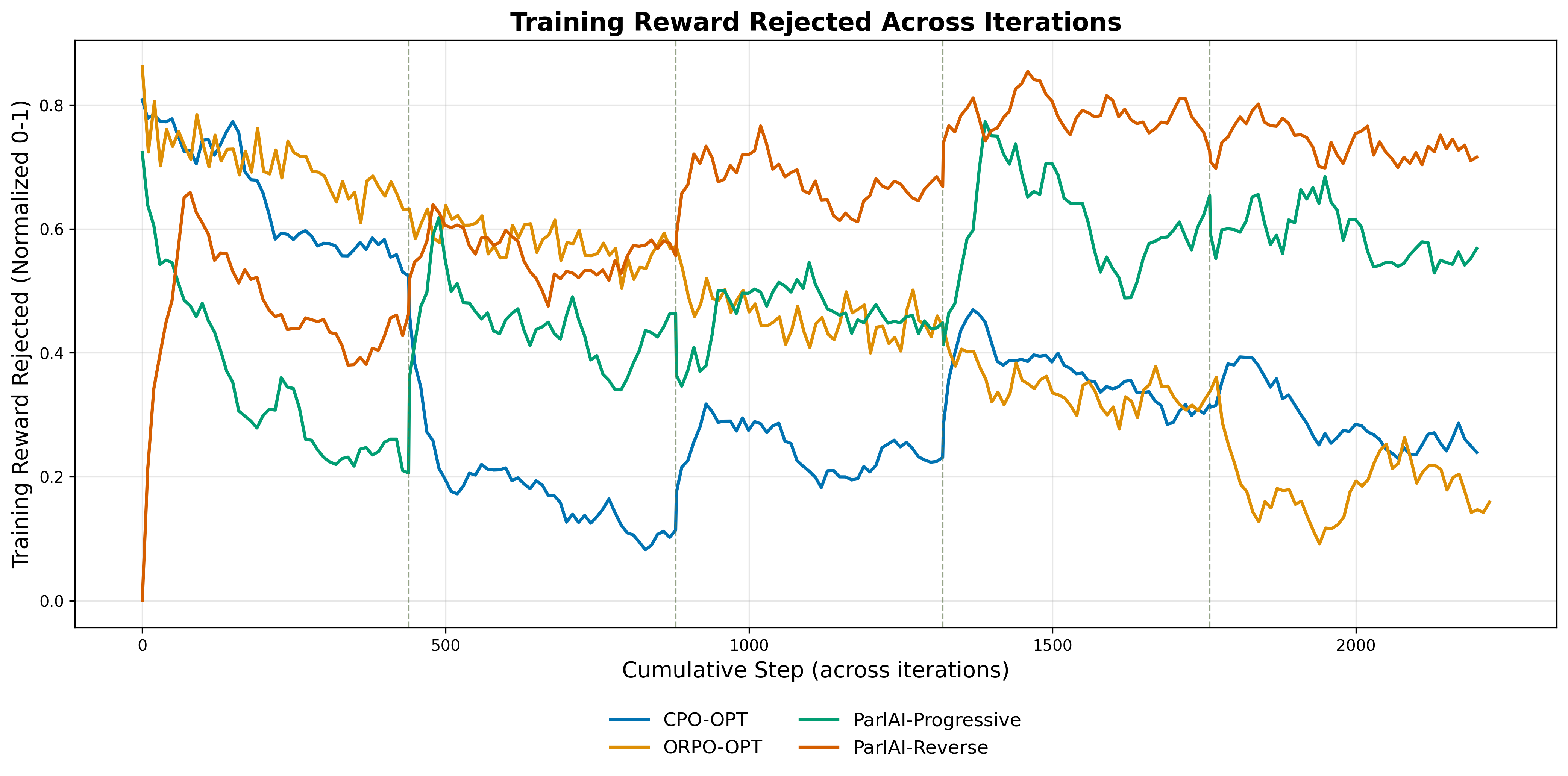}
    \caption{\name Training Rejection for CPO/ORPO Models (Experiment 1) and CEFR Models (Experiment 2)}
    \label{fig:reward_rejected_1}
\end{figure*}

\begin{figure*}[h!]
    \centering
    \includegraphics[width=\textwidth]{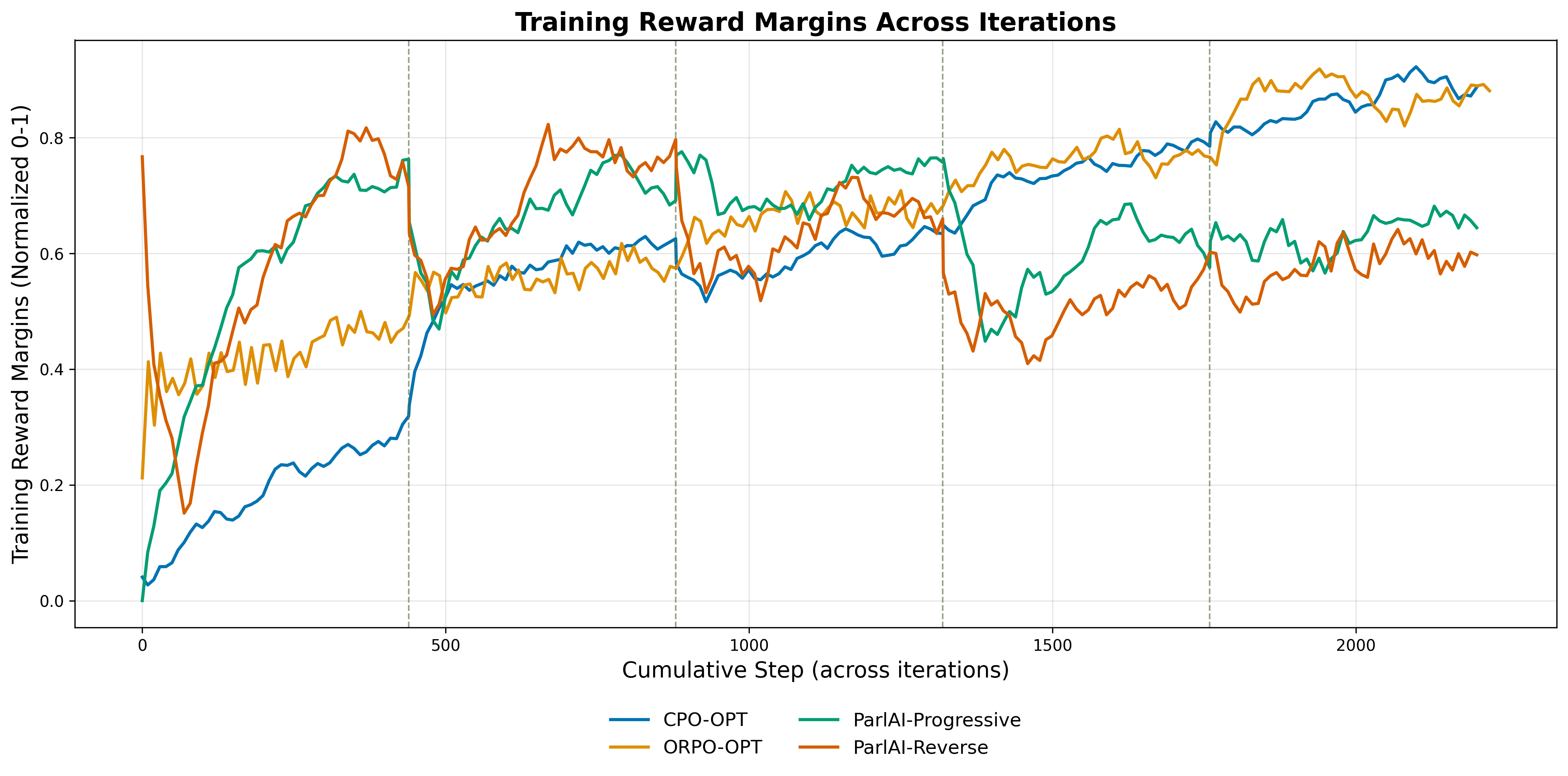}
    \caption{\name Training Reward Margin for CPO/ORPO Models (Experiment 1) and CEFR Models (Experiment 2)}
    \label{fig:reward_margins_1}
\end{figure*}
\newpage 

\subsection{Progressive CEFR Model Training Reward Dynamics Across Iterations and Reward Types (Experiment 2)}

\begin{figure*}[h!]
    \centering
    \includegraphics[width=\textwidth]{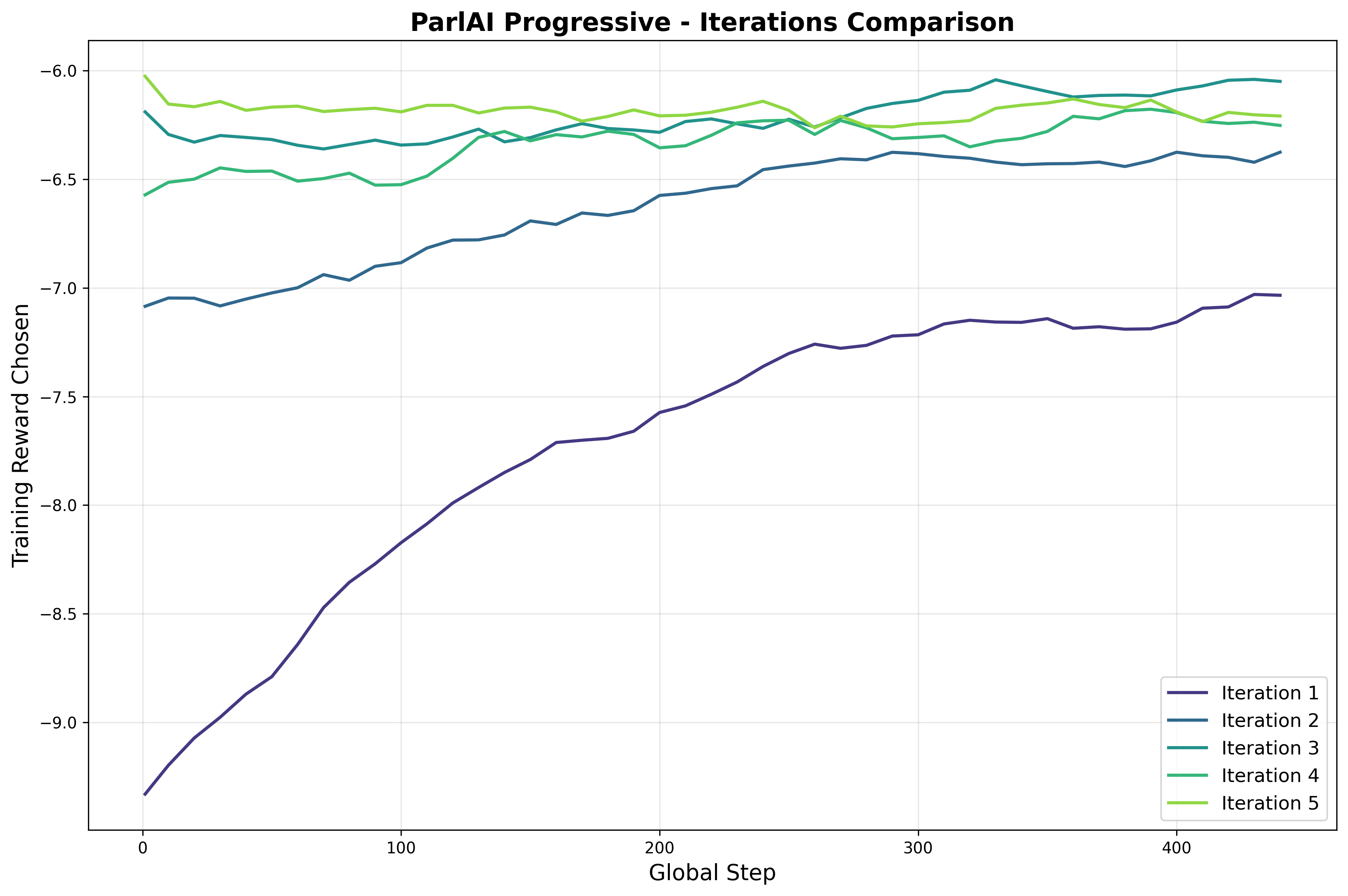}
    \caption{\name Training Reward Chosen of Progressive CEFR model across iterations}
    \label{fig:reward_chosen_parlai_progressive}
\end{figure*}

\begin{figure*}[h!]
    \centering
    \includegraphics[width=\textwidth]{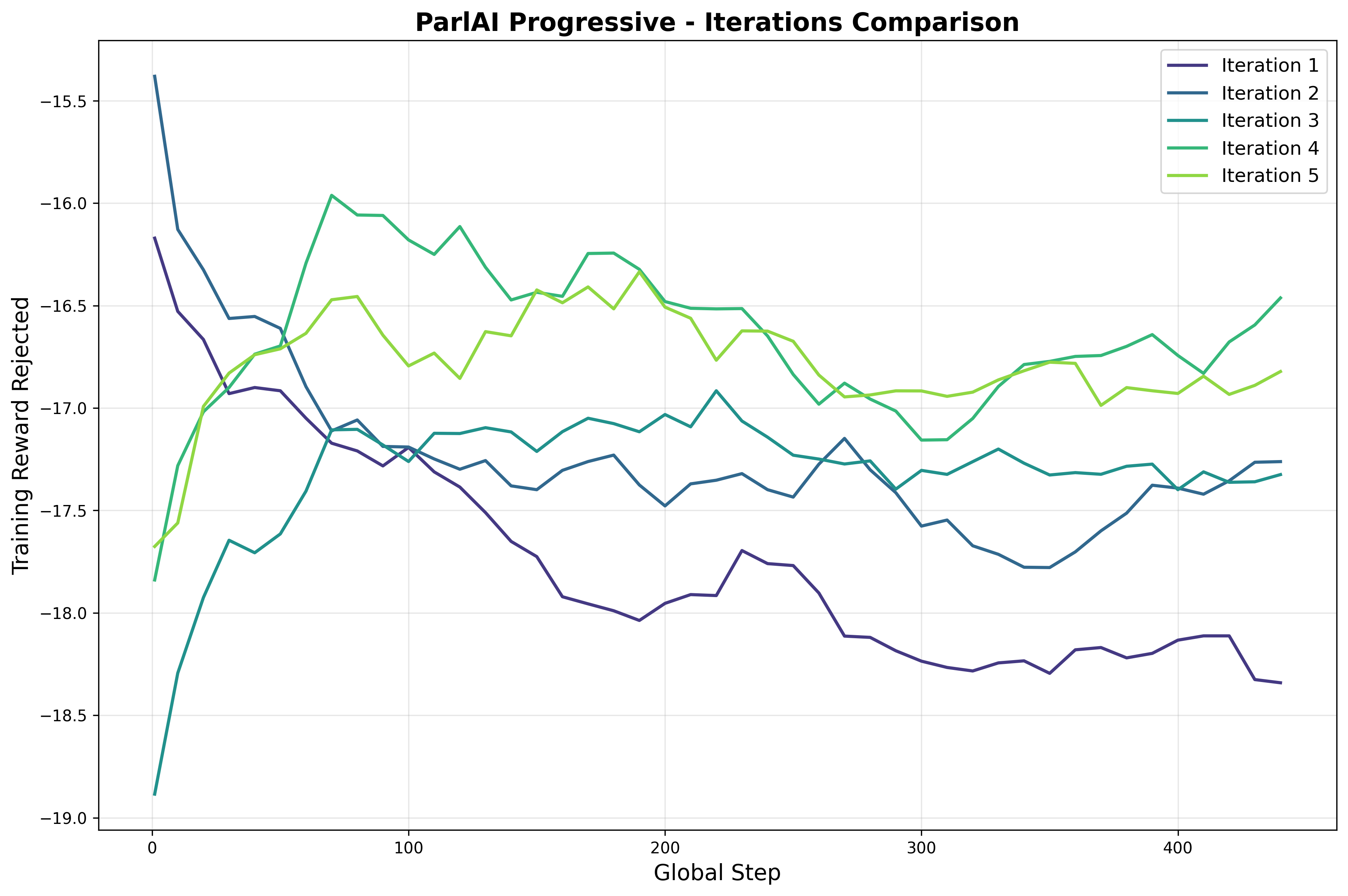}
    \caption{\name Training Reward Rejected of Progressive CEFR model across iterations}
    \label{fig:reward_rejected_parlai_progressive}
\end{figure*}
 
\begin{figure*}[h!]
    \centering
    \includegraphics[width=\textwidth]{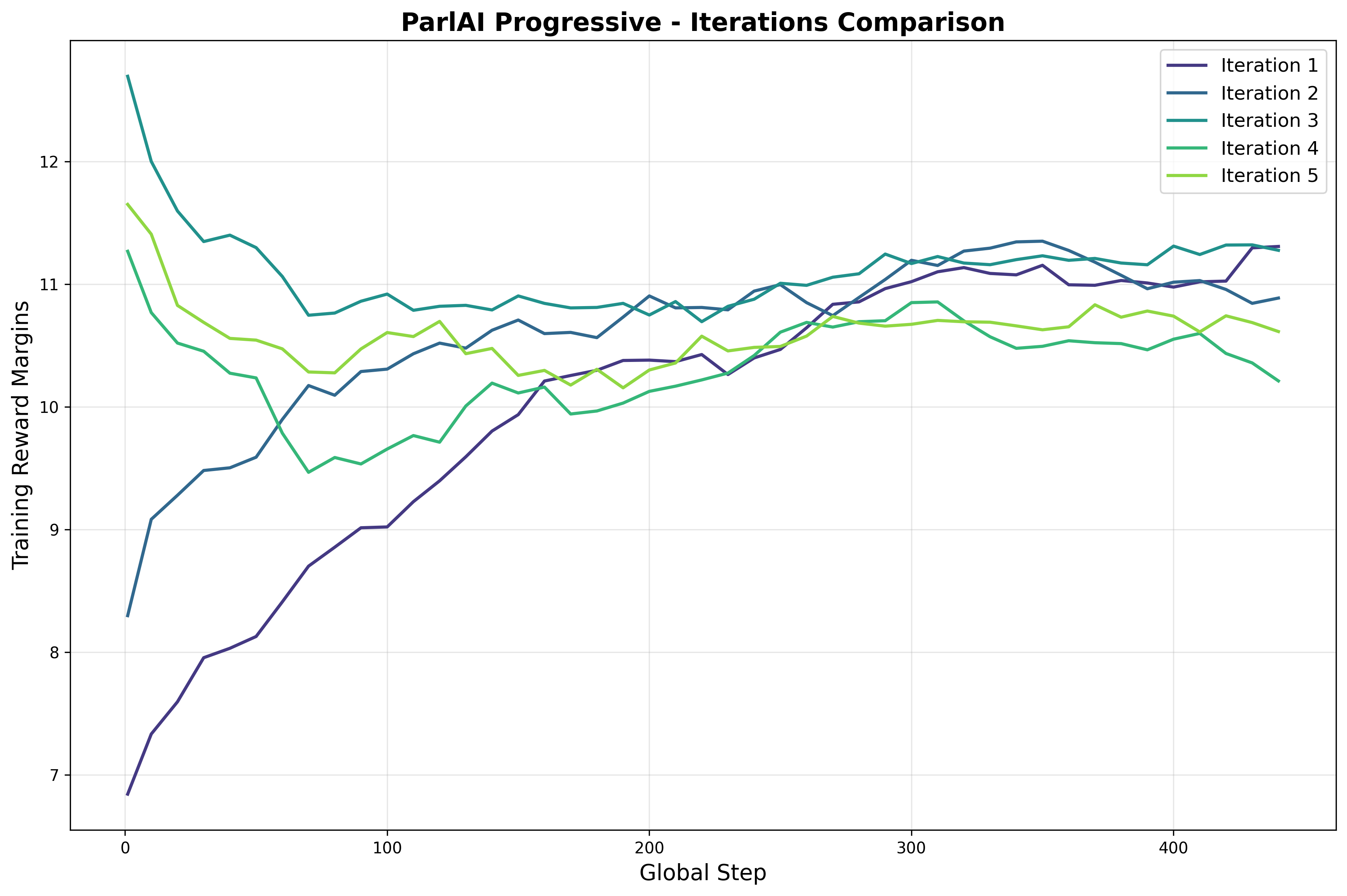}
    \caption{\name Training Reward Margins of Progressive CEFR model across iterations}
    \label{fig:reward_margins_parlai_progressive}
\end{figure*}

\clearpage 
\subsection{Reverse CEFR Model Training Reward Dynamics Across Iterations and Reward Types (Experiment 2)}

\begin{figure*}[h!]
    \centering
    \includegraphics[width=\textwidth]{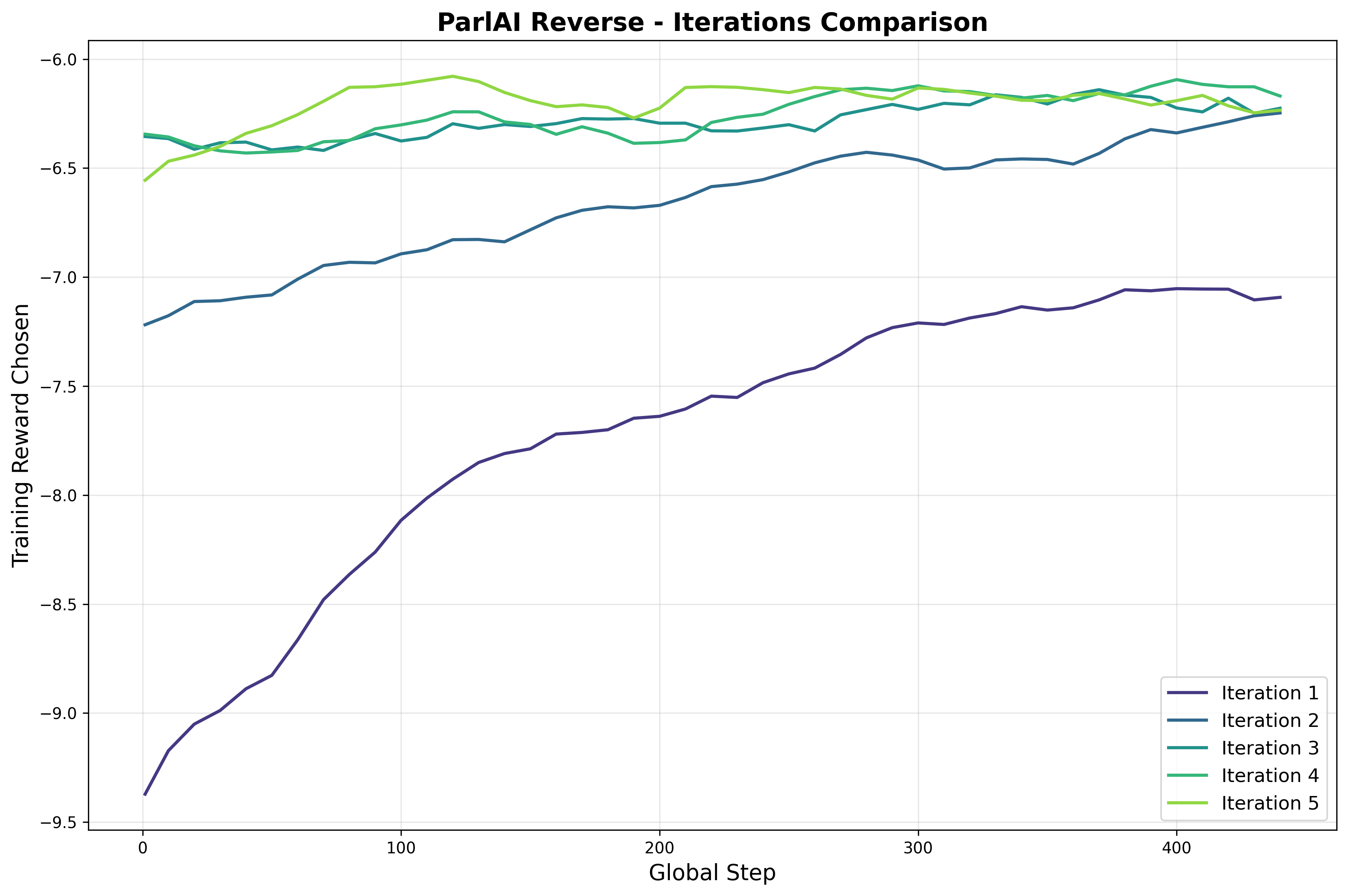}
    \caption{\name Training Reward Chosen of Reverse CEFR model across iterations}
    \label{fig:reward_chosen_parlai_reverse}
\end{figure*}

\begin{figure*}[h!]
    \centering
    \includegraphics[width=\textwidth]{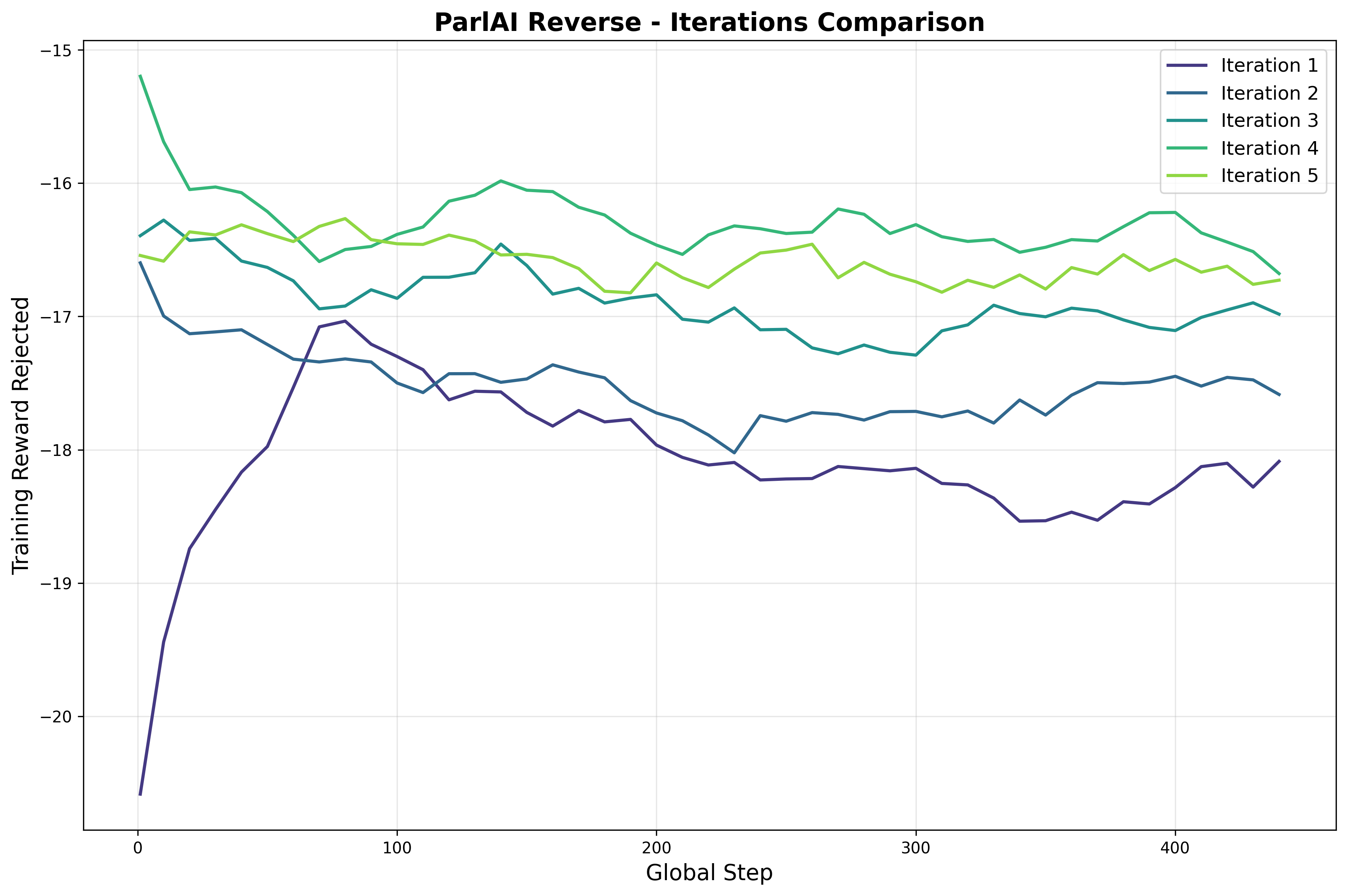}
    \caption{\name Training Reward Rejected of reverse CEFR model across iterations}
    \label{fig:reward_rejected_parlai_reverse}
\end{figure*}

\begin{figure*}[h!]
    \centering
    \includegraphics[width=\textwidth]{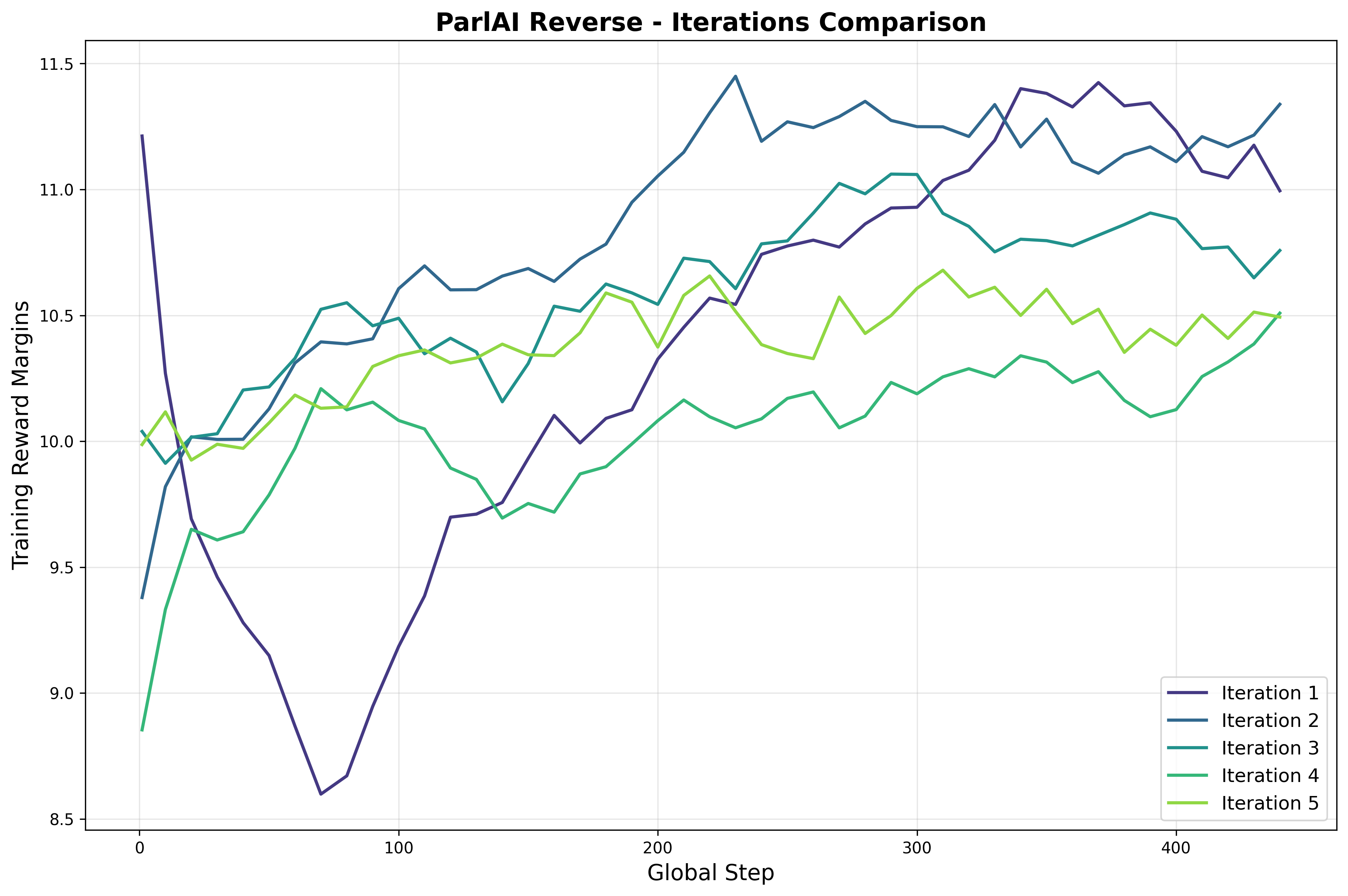}
    \caption{\name Training Reward Margins of reverse CEFR model across iterations}
    \label{fig:reward_margins_parlai_reverse}
\end{figure*}

\end{document}